\documentclass[12pt]{article}

\usepackage[brazil]{babel}

\usepackage[utf8]{inputenc}

\usepackage{subfiles}

\usepackage[alf]{abntex2cite}

\usepackage{amssymb}

\usepackage{subcaption}
\usepackage{csvsimple}
\usepackage{booktabs}

\usepackage{tikz,pgfplots}

\newcommand{\tsp}{\mbox{\rm TSP}}

\newcommand{\vrp}{\mbox{VRP}}
\newcommand{\depo}{\pi}
\newcommand{\nbb}{\mathbb{N}}
\newcommand{\qbb}{\mathbb{Q}}
\newcommand{\sol}{\mathcal{S}}
\newcommand{\rota}{\mathcal{R}}
\newcommand{\rotamax}{\rota_{max}}
\newcommand{\particao}[1]{Particao(#1)}
\newcommand{\postvrp}{PostVRP}
\newcommand{\compri}[1]{W(#1)}

\newtheorem{definicao}{Definição}

\usepackage[portuguese,ruled]{algorithm2e}

\begin{document}


\title{
Algoritmos Genéticos Aplicado ao Problema de Roteamento de Veículos
}

\author{Felipe F. Müller \and Luis A. A. Meira}

\maketitle

\renewcommand{\abstractname}{Abstract}
\begin{abstract}
Routing problems are often faced by companies who serve costumers through vehicles. Such problems have a challenging structure to optimize, despite the recent advances in combinatorial optimization. The goal of this project is to study and propose optimization algorithms to the vehicle routing problems (VRP). Focus will be on the problem variant in which the length of the route is restricted by a constant. A real problem will be tackled: optimization of postmen routes. Such problem was modeled as {multi-objective} in a roadmap  with  25 vehicles and {30,000 deliveries} per day  .
\end{abstract}

\renewcommand{\abstractname}{Resumo}

\begin{abstract}
Problemas de roteamento são frequentemente enfrentados por empresas que atendem clientes através de veículos. Tais problemas possuem uma estrutura desafiadora de otimizar, apesar dos recentes avanços na otimização combinatória. O objetivo deste mestrado é estudar e propor algoritmos de otimização para o problema de roteamento de veículos (VRP). O foco estará na variante do problema na qual o comprimento da rota é limitado por uma constante. Um problema real será abordado: otimização de rotas de carteiros. Tal problema foi modelado como {multi-objetivo} em uma malha viária com 25 veículos e até {30.000 entregas} por dia.
\end{abstract}

\section{Introdução}
\label{cap:introducao}

\subsection{Descrição Formal do Problema}
\label{sec:definicao}

Considere um grafo ponderado $G(V,E)$ e uma função custo  $w':E\rightarrow \qbb^+$. O grafo pode ser orientado ou não. Existe um  vértice especial $\depo\in V$ chamado depósito. 
Existem variantes do VRP com múltiplos depósitos, as quais não serão tratadas nesta seção.
O conjunto de clientes é dado por $C=V\setminus\{\depo\}$. O número de clientes é dado por $n=|C|$. O conjunto de clientes pode ser representado por $C=\{c_1,\ldots,c_n\}$.
Existe um valor $k\in \nbb$
que representa o número de veículos. O valor de $k$ pode ser uma constante parte da entrada ou pode ser uma variável definida no processo de otimização.
Seja $w(u,v,G)$ o custo menor caminho entre os vértices $u$ e $v$ no grafo $G$. Se $G$ estiver claro no contexto, pode-se usar $w(u,v)$ para representar $w(u,v,G)$.

Usaremos uma sequência de elementos para representar uma solução do problema, conforme descrito abaixo:

$$\sol(C,k) = (c_1,\ldots,c_n,\pi,\ldots,\pi).$$

Esta sequencia é montada da seguinte maneira. Primeiro, todos os clientes são  inseridos em $\sol$. Após isso, o vértice depósito 
é inseriodo $k-1$ vezes. Se o conjunto de clientes $C$ e o número de veículos $k$ estiverem claros no contexto, poderemos 
usar $\sol$ para represnetar $\sol(C,k)$.

\begin{figure}[htb]
\begin{center}
\includegraphics[width=0.6\textwidth]{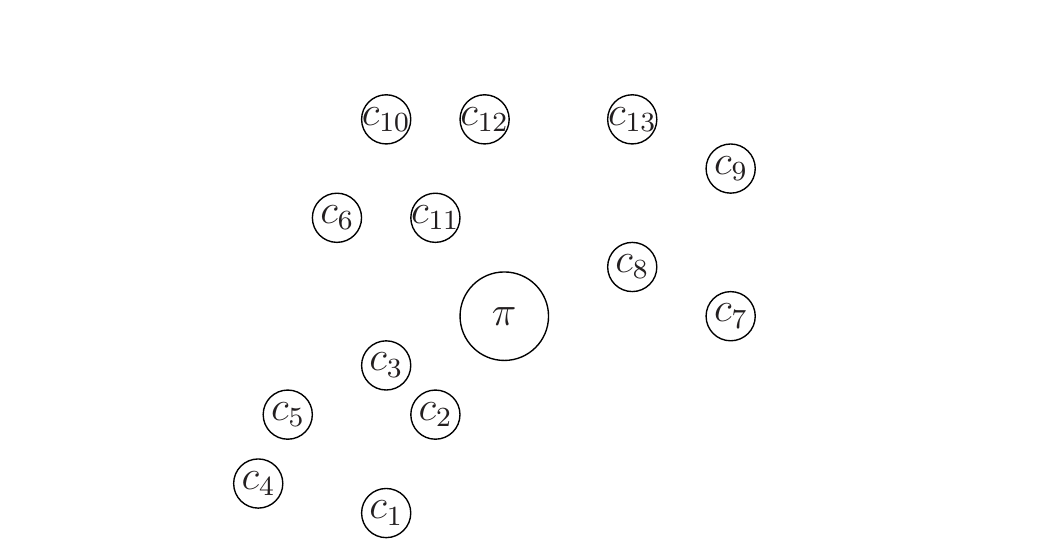}
\end{center}
\caption{Exemplo de Instância do VRP}
\label{fig:instance}
\end{figure}

Cada permutação de $\sol$ representa uma solução do VRP. 
Para exemplificar, considere o grafo ilustrado em \ref{fig:instance}.

Considere os vértices descritos na figura~\ref{fig:instance}. Considere o número de veículos igual a 3, ou seja $k=3$. A sequência $\sol(C,3)$ neste caso seria:
$\sol(C,3)=(c_1,\ldots,c_{13},\pi,\pi).$

\begin{figure}[htb]
\begin{center}
\includegraphics[width=0.4\textwidth]{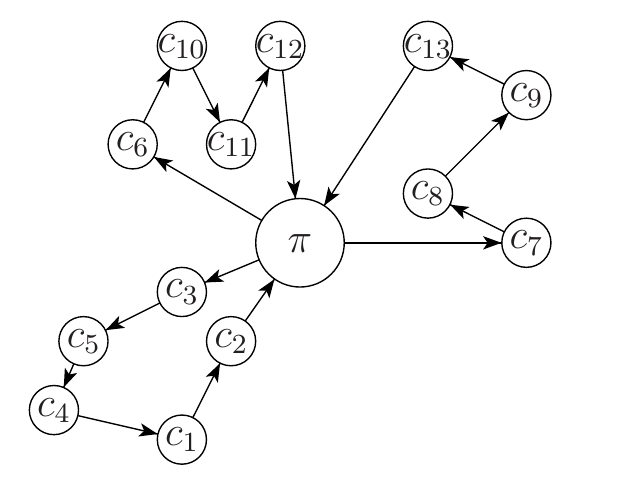}
\end{center}
\caption{Solução $\sol'=(c_3,c_5,c_4,c_1,c_2,\mathbf{\pi},c_6,c_{10},c_{11},c_{12},\mathbf{\pi},c_7,c_8,c_9,c_{13})$. }
\label{fig:solinst}
\end{figure}

Cada permutação de $\sol(C,3)$ representa uma solução para o VRP. Por exemplo, a permutação
$\sol'=(c_3,c_5,c_4,c_1,c_2,\pi,c_6,c_{10},c_{11},c_{12},\pi,c_7,c_8,c_9,c_{13})$ representa a solução descrita na figura~\ref{fig:solinst}.

Toda rota inicia e termina no depósito.
A solução $\sol'$ representa uma partição dos clientes em 3 rotas: $\rota_1=(c_3,c_5,c_4,c_1,c_2)$, 
$\rota_2=(c_6,c_{10},c_{11},c_{12})$ e $\rota_3=(c_7,c_8,c_9,c_{13})$. Mais formalmente, uma solução pode ser particionada em diversas rotas:
$$\displaystyle\particao{\sol}=\left(\rota_1,\ldots,\rota_{k'}\right),$$ onde a partição da solução é feita no depósito $\depo$.
No exemplo da figura~\ref{fig:solinst} temos que: 
$\particao{\sol'}=
\left(\left \{c_3,c_5,c_4,c_1,c_2\right\},\left\{c_6,c_{10},c_{11},c_{12}\right\},\left\{c_7,c_8,c_9,c_{13}\right\}\right). 
$
Por definição, rotas vazias não fazem parte de  $\particao{\sol}$. Assim $\particao{1,2,\depo,\depo,3,4}$ é $\{(1,2),(3,4)\}$ e não $\{(1,2),(),(3,4)\}$, 
 portanto $k'\leq k$.

O comprimento de uma rota $\rota=(r_1,\ldots,r_m)$ é dado por:
$$\compri{\rota}=w(\depo,r_1)+w(r_m,\depo)+\sum_{i=1}^{m-1}w(r_i,r_{i+1}).$$

O comprimento de uma solução $\sol=(s_1,\ldots,s_m)$ é calculado de maneira análoga:
$$\compri{\sol}=w(\depo,s_1)+w(s_m,\depo)+\sum_{i=1}^{m-1}w(s_i,s_{i+1}).$$

O número de veículos usados em uma solução é igual ao número de rotas não vazias $|\particao{\sol}|.$
Se o número de veícuos é $k$ e não são permitidas rotas vazias, temos a restrição $|\particao{\sol}|=k$.
Se o número de veículos é no máximo $k$, ou se rotas vazias são permitidas, nós temos $|\particao{\sol}|\leq k$.
Se o número de veículos $k$ não é parte da entrada, o domínio pode ser representado pela permutação da sequência
$\sol(C,n)$. Neste caso, o número de veículos é definido durante o processo de otimização.

Dada uma solução viável, precisamos calcular o seu custo. O valor mais tradicional a ser minimizado é o comprimento de uma solução, $$f_1(\sol)=\compri{\sol}.$$ Outra função objetivo consiste em minimizar o número de veículos.

$$f_2(\sol)=|\particao{\sol}|.$$

Finalmente, é necessário gerar rotas justas, que não penalizem um carteiro em relação a outro. 
Uma maneira de representar a justiça é por meio do cálculo da variância entre o comprimento das rotas de uma dada solução:

$$\displaystyle f_3(\sol)=\sqrt{\frac{\displaystyle\sum_{\rota\in\particao{\sol}}\left(\compri{\rota}-\overline{\compri{\rota}}\right)^2}{|\particao{\sol}|-1}}.$$



Consideraremos uma variante do problema onde o tamanho da rota é limitada. Este problema pode modelar veículos que precisam reabastecer no depósito, como um helicóptero por exemplo. Também pode modelar situações trabalhistas, onde o condutor de um veículo tem um limite de disponibilidade de tempo. Considere que existe um limite  $\rotamax$ para o comprimento da rota.  Desta forma, toda rota $\rota$ precisa respeitar a seguinte restrição para ser considerada viável:
$$W(\rota)\leq \rotamax.$$

%
%

Abaixo temos a definição do problema de interesse, que será chamado de \postvrp, ou seja, VRP no contexto dos correios \emph{(Post Office)}.

\begin{definicao}[\postvrp] Dado um grafo ponderado $G(V,E)$, uma constante $k$ representando o número máximo de veículos, um vértice especial 
$\pi\in V$ e o comprimento máximo de uma rota $\rotamax$. Seja $C\gets V\setminus\{\pi\}$. Considere a sequencia $\sol(C,k)$. Seja  $P$ o conjunto de  todas as permutações de $\sol(C,k)$. Encontre as permutações  $\sol^*\subset P$ viáveis em relação a $\rotamax$ e que minimizem 
as funções objetivos  $f_1(\sol)$, $f_2(\sol)$ e $f_3(\sol)$.
\end{definicao}

Nas próximas subseções iremos descrever as técnicas de otimização comumente utilizadas para
resolver VRP. A seguir os métodos descritos no livro Bio-algs.

\subsection{Métodos Diretos}
\label{sec:diretos}

Diferente de meta-heurísticas, as quais são suficientemente genéricas para serem aplicadas a uma grande variedade de problemas, os métodos diretos são projetados de forma mais focada para serem aplicados a um problema específico.

Também conhecidos como heurísticas clássicas, os métodos diretos para o \vrp~foram desenvolvidos principalmente entre 1960 a 1990. Capazes de produzir boas soluções em tempos computacionais modestos, eles executam uma exploração mais limitada do espaço de busca e podem ser facilmente adaptados à restrições encontradas no mundo real.

Heurísticas clássicas podem ser divididas em três categorias, as quais serão abordadas nas subseções \ref{subsec:construtivo}, \ref{subsec:2fases} e \ref{subsec:incremental}. São elas: métodos construtivos, que vão construindo soluções viáveis aos poucos; métodos em duas fases, que dividem o problema em agrupamento de vértices e construção de rotas; e métodos incrementais, que tentam melhorar soluções iniciais através de trocas de vértices ou arestas.

\subsubsection{Construtivos}
\label{subsec:construtivo}

O método construtivo mais conhecido para o VRP é provavelmente o \emph{Clark and Wright Savings Algorithm} (CWS). O algoritmo se baseia na união de duas rotas de modo que a rota resultante gere uma redução de custo, chamada ganho ou \emph{saving}. Particularmente indicado para problemas sem um número fixo de veículos, o CWS pode ser implementado em duas versões: \textit{(i)} sequencial, onde é construída uma rota de cada vez e \textit{(ii)} paralela, onde várias rotas são criadas ao mesmo tempo.

O Algoritmo~\ref{algo:cws} é o CWS paralelo. Tal algoritmo
inicia com $n$ rotas, cada uma partindo do depósito, visitando um único ponto de entrega e retornando ao depósito. Desta forma, o custo da solução começa bastante alto.

\sloppy

Para cada par de vértices há um valor associado que representa o ganho ao se juntar duas rotas. Para calcular este ganho, considere dois pontos de entrega $d_1$ e $d_2$ e considere duas rotas $r_1=(depot,d_1,depot)$ e
$r_2=(depot,d_2,depot)$. Ao se unir tais rotas teremos $r_3=(depot,d_1,d_2,depot)$. Sendo assim, 
o ganho é dado por: 
\begin{eqnarray}s(d_1,d_2)=w(depot,d_1)+w(depot,d_2)-w(d_1,d_2).\label{saving}\end{eqnarray}
Note que 
$s(d_1,d_2)$ é igual ao $custo(r_1)+custo(r_2)-custo(r_3)$.
Uma vantagem do CWS é que o valor $s(d_1,d_2)=w(depot,d_1)+w(depot,d_2)-w(d_1,d_2)$ é fixo e pode ser calculado uma única vez no inicio do algoritmo. 

Um detalhe importante, é que o conceito de \emph{saving} envolvendo rotas maiores, permite a união apenas nos extremos. Seja $prefixo$ e $sufixo$ dois conjuntos quaisquer de entregas,
considere duas rotas $r_1=(depot,prefixo,d_1,depot)$ e
$r_2=(depot,d_2,sufixo,depot)$. Ao se unir tais rotas teremos $r_3=(depot,prefixo,d_1,d_2,sufixo,depot)$. O ganho será o mesmo mostrado na equação~\ref{saving}.

Em suma, as rotas iniciam com tamanho um e vão sendo incrementadas. A cada iteração, um vértices ou uma rota é concatenada a outra. Note que a união só é feita se respeitar à restrições, como as de capacidade por exemplo.

Na versão sequencial, existe uma única rota que é incrementada em um vértice a cada iteração. A escolha é sempre do vértice que ofereça o maior ganho. Ao atingir a capacidade do veículo, a rota atual é fechada e uma nova rota é inciada.

Já na versão paralela, várias rotas são criadas ao mesmo tempo. Sempre são concatenadas as duas rotas tais que seus extremos $s_1$ e $s_2$ possuam o máximo \emph{saving} $s[s_1][s_2]$, considerando todos os pares de extremos possíveis.

Esta heurística tem uma boa qualidade no inicio das rotas, colocando sempre um novo elemento promissor. Entretanto as rotas tendem a crescer deixando pontos para trás. Tais pontos acabam depois sendo cobertos por meio de rotas ruins, de alto custo.


\begin{algorithm}[htb!]
\LinesNumbered
\Entrada{Um grafo $G(V,E)$ com pesos $w:E\rightarrow \mathbb{R}^+$, um inteiro $k\in \mathbb{N}$,
um depósito $\pi\in V$ e o comprimento máximo de uma rota $\mathcal{R}_max\in \mathbb{R}^+$. Seja $C=V\setminus\{\pi\}$. 
}
\Saida{Uma permutação $\mathcal{S}(C,k)\in P$}
Calcule $s[u][v]$ para todo $u\neq v$ em $V$.\\
Seja $S=\{s[u][v]~|~(u,v)\in V\times V, u\neq v\} $\\
Seja $S'$ uma ordenação decrescente de $S$.\\
Seja $Sol$ um conjunto de rotas, inicialmente $Sol=C$, ou seja cada vértice é uma rota com um único elemento\\
\For{$i=1,i\leq|S'|,i++$,}{
Considere o saving $s'_i\in S'$ e seus vertices $u$ e $v$.\\
\If{$u$ for o extremo de uma rota $R'\in Sol$ e $v$ for o extremo de outra
rota $R''\in Sol$}{
\If{custo $R'\cup R''$ for viável }{
$Sol\gets Sol\setminus\{R',R''\}$\\
$Sol\gets Sol\cup\{R'R''\}$
}
}
}

\Return{$Sol$}\;
\caption{Algoritmo CWS paralelo para o CVRP}
\label{algo:cws}
\end{algorithm}
\begin{figure}[h]
  \centering
  \begin{subfigure}{.3\textwidth}
  \includegraphics[scale=0.5]{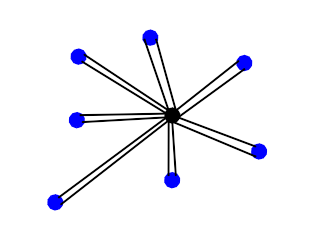}
  \caption{Cada vértice começa visitado por uma rota.}
  \end{subfigure}
  \begin{subfigure}{.3\textwidth}
  \includegraphics[scale=0.5]{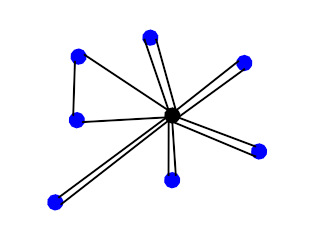}
  \caption{Duas rotas são unidas pela concatenação dos vértices com maior \emph{saving}.}
  \end{subfigure}
  \begin{subfigure}{.3\textwidth}
  \includegraphics[scale=0.5]{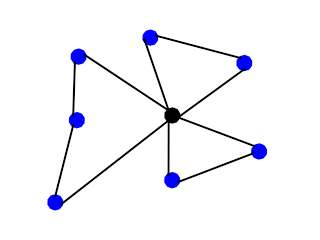}
  \caption{Solução resultante das uniões respeitando os limites de cada veículo.}
  \end{subfigure}
  
  \caption{Ação do CWS no grafo.}
  \label{fig:cws}
\end{figure}

O livro \cite{toth2001vehicle} ainda cita mais duas heurísticas construtivas, são elas o \emph{Matching-Based Savings Algorithm}~\cite{MBS_Kemal} e a \emph{Sequential Insertion Heuristic}~\cite{Insertion}.

\subsubsection{Duas Fases}
\label{subsec:2fases}

\begin{figure}[h]
  \centering
  \begin{subfigure}{.3\textwidth}
  \includegraphics[scale=0.5]{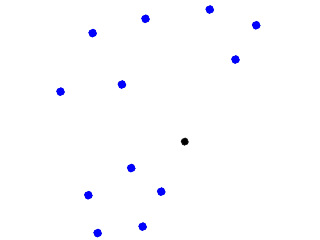}
  \caption{Grafo}
  \end{subfigure}
  \begin{subfigure}{.3\textwidth}
  \includegraphics[scale=0.5]{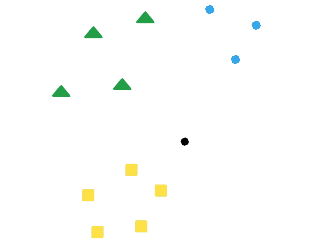}
  \caption{Formação dos \emph{clusters}}
  \end{subfigure}
  \begin{subfigure}{.3\textwidth}
  \includegraphics[scale=0.5]{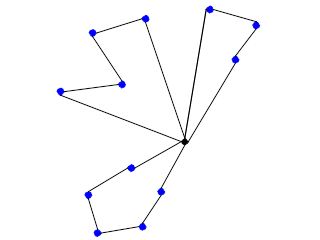}
  \caption{Roteamento}
  \end{subfigure}
  
  \caption{VRP em duas fases: \emph{cluster-first, route-second}.}
  \label{fig:clusterfirst}
\end{figure}

\begin{figure}[h]
  \centering
  \begin{subfigure}{.3\textwidth}
  \includegraphics[scale=0.5]{pontos.png}
  \caption{Grafo}
  \end{subfigure}
  \begin{subfigure}{.3\textwidth}
  \includegraphics[scale=0.5]{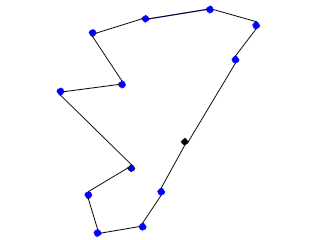}
  \caption{Roteamento}
  \end{subfigure}
  \begin{subfigure}{.3\textwidth}
  \includegraphics[scale=0.5]{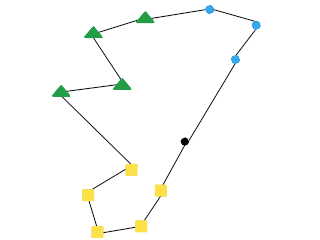}
  \caption{Formação dos \emph{clusters}}
  \end{subfigure}
  \begin{subfigure}{.3\textwidth}
  \includegraphics[scale=0.5]{rotas.png}
  \caption{Quebra em rotas viáveis}
  \end{subfigure}
  
  \caption{VRP em duas fases: \emph{route-first, cluster-second}.}
  \label{fig:routefirst}
\end{figure}

As heurísticas em duas fases são divididas em duas abordagens, \textit{cluster-first, route-second} ou \textit{route-first, cluster-second}. Usando \textit{cluster-first, route-second}, um conjunto de vértices é particionado em subconjuntos que então são roteados separadamente. Já usando \textit{route-first, cluster-second}, uma grande rota é traçada para o conjunto completo de vértices e depois uma técnica de agrupamento é aplicada para a quebra do resultado em várias rotas menores. As figuras \ref{fig:clusterfirst} e \ref{fig:routefirst} ilustram estas duas abordagens.

\begin{figure}[h]
  \centering
  \begin{subfigure}{.3\textwidth}
  \includegraphics[scale=0.4]{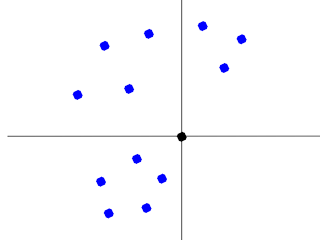}
  \caption{Grafo no plano cartesiano.}
  \end{subfigure}
  \begin{subfigure}{.3\textwidth}
  \includegraphics[scale=0.4]{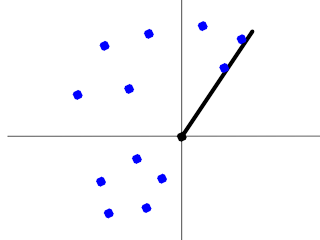}
  \caption{Raio traçado a partir do depósito.}
  \end{subfigure}
  \begin{subfigure}{.3\textwidth}
  \includegraphics[scale=0.4]{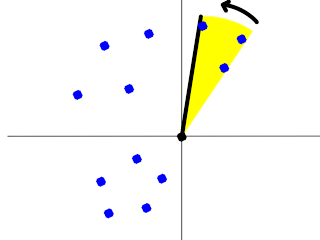}
  \caption{Area varrida até que os vértices selecionados atingissem o limite do veículo.}
  \end{subfigure}
  \begin{subfigure}{.3\textwidth}
  \includegraphics[scale=0.4]{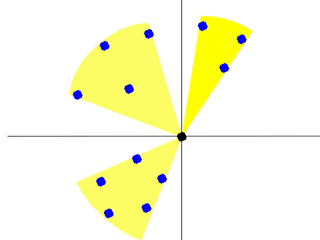}
  \caption{Repetição dos passos b e c até todos os vértices serem varridos.}
  \end{subfigure}
  \begin{subfigure}{.3\textwidth}
  \includegraphics[scale=0.4]{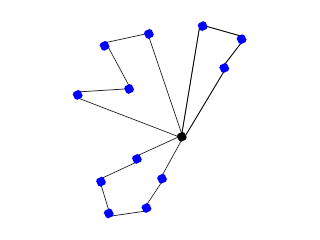}
  \caption{Roteamento dos \emph{clusters} gerados.}
  \end{subfigure}
  
  \caption{Execução do \emph{Sweep Algorithm}.}
  \label{fig:sweep}
\end{figure}

Um dos algoritmos mais simples do tipo \emph{Cluster-first} é o \emph{Sweep Algorithm}~\cite{sweepAlg}. Em tal algoritmo, cada vértice a ser visitado possui coordenadas polares. Estes vértices são varridos por um raio traçado a partir do depósito. Esta varredura vai então atribuindo os vértices a um veículo em ordem crescente de ângulo até que o limite de capacidade do veículo seja atingido. O processo se repete até que todos os vértices pertençam a um veículo. Finalmente, cada rota gerada é otimizada individualmente como um TSP. A figura~\ref{fig:sweep} ilustra o procedimento aqui descrito.

Outros algoritmos \emph{Cluster-first} citados em \cite{toth2001vehicle} são o \emph{Fisher and Jaikumar Algorithm} (FJA)~\cite{FJA} e o \emph{Petal Algorithm}~\cite{petal}.

Em relação à abordagem \emph{Route-first}, ela foi proposta pela primeira vez por Beasley~\cite{routefirst}. Como a segunda etapa da abordagem pode ser vista como um problema de caminho minimo em um grafo acíclico, é possível resolve-la com o algoritmo de Dijkstra~\cite{Dijkstra:1959:NTP:2722880.2722945}, por exemplo. Considerando $custo(O,i)+custo(O,j)+l(i,j)$ usado para calcular o $custo(i,j)$ no problema de caminho minimo, o valor de $l(i,j)$ corresponde ao custo de viajar de i até j no percurso gerado na etapa de roteamento.

Segundo Toth~\cite{toth2001vehicle}, não há estudos demonstrando métodos \emph{Route-first} que sejam competitivos à outras técnicas.

\subsubsection{Incrementais}
\label{subsec:incremental}

Métodos incrementais são utilizados para melhorar soluções já existentes. Caso cada rota do VRP seja trabalhada individualmente, qualquer técnica para TSP pode ser empregada. Caso contrário, também existem alguns métodos que trabalham com múltiplas rotas.

Entre as abordagens para rotas individuais, podemos citar três:

\begin{itemize}
\item Movimento de vértice: um ou mais vértices são movidos para outra posição;
\item Troca de vértices: um par de vértices trocam de posição;
\item Troca de arestas: duas ou mais arestas trocam de posição.
\end{itemize}

A troca de arestas é também conhecida como $\lambda$-opt\cite{lin-opt}. O algoritmo remove $\lambda$ arestas da solução e tenta reconectar os segmentos resultantes de modo a reduzir o custo da rota. Tal processo se repete até que um ótimo local seja atingido. O algoritmo \ref{alg:2opt} descreve uma implementação do 2-opt.

\begin{algorithm}[htb]
\While{houve melhora}{
	\For{cada par de arestas não consecutivas}{
    	encontrar o par cuja troca de posição melhor reduz o custo da solução;
        }
    arestas são trocados de posição;
    }
    \caption{Implementação do 2-opt.}
\label{alg:2opt}
\end{algorithm}

Executando em tempo $O(n^\lambda)$, suas variantes mais comuns são o 2-opt e 3-opt, troca de 2 e 3 arestas, respectivamente. Johnson and McGeoch~\cite{johnson1997traveling} analisaram o desempenho de algoritmos incrementais para o TSP e concluíram que o melhor era uma implementação do $\lambda$-opt com $\lambda$ dinâmico~\cite{lin-opt-melhor}.


Entre as abordagens que trabalham com múltiplas rotas, Thompson e  Psaraftis~\cite{ThompsonCyclic1993} descreveram uma técnica geral conhecida como \emph{$b$-cyclic, $k$-transfer exchanges}, na qual $k$ clientes de uma rota são movidos para a rota seguinte em um ciclo de permutações com $b$ rotas. Dentro desta premissa, Breedam~\cite{Breedam94} classifica 4 tipos de operações: \emph{String cross} (SC), na qual duas \emph{strings} de vértices são trocadas cruzando arestas entre duas rotas; \emph{String exchange} (SE), na qual duas \emph{strings} com um número máximo de $k$ vértices são trocadas entre duas rotas; \emph{String relocation} (SR), na qual uma \emph{string} com um número máximo de $k$ vértices é transferida a outra rota; e \emph{String mix} (SM), na qual é feita a melhor operação entre SE e SR.


\subsection{Algoritmos Genéticos e Bio-inspirados}
\label{sec:geneticos}

Algoritmos bio-inspirados são soluções computacionais desenvolvidas a partir de comportamentos observados na natureza~\cite{HorowitzS78}. 
A seleção natural, a busca de comida por formigas e as sinapses cerebrais são alguns exemplos de ``mecanismos'' naturais relativamente bem conhecidos pela biologia que funcionam  bem 
para seus respectivos propósitos e portanto servem de inspiração para heurísticas.

A implementação de tais algoritmos é obviamente uma simplificação da realidade
 que exige uma boa adaptação ao caso em questão e não garante resultados ótimos.
 
 
 Algortimos genéticos/evolutivos, por exemplo, possuem as seguintes etapas~\cite{back1996evolutionary}:

\begin{enumerate}
\item Gerar uma população inicial de soluções;
\item Filtrar as soluções através de ``pressões ambientais'';
\item Cruzar as soluções sobreviventes;
\item Introduzir aleatoriamente um ``fator de mutação'';
\item Obter a nova geração de soluções e
\item Repetir o processo a partir da etapa 2.
\end{enumerate}

No caso de algoritmos genéticos, cada permutação da sequencia  $\sol(C,k)$ pode representar um indivíduo da população. Existem diversas possibilidade de cruzamento entre dois indivíduos. A regra consiste em gerar um vetor $\sol^3 $ válido a partir de dois vetores  $\sol^1 $  e  $\sol^2 $. 

No capítulo I do livro~\cite{pereira2008bio}, foi descrito uma metodologia para cruzamento de soluções do TSP adaptada com sucesso para o VRP.
Tal técnica chama-se \emph{EAX crossover} (\emph{Edge Assembly Crossover}).

  \begin{figure}[ht]\centerline{
\includegraphics[width=0.8\textwidth]{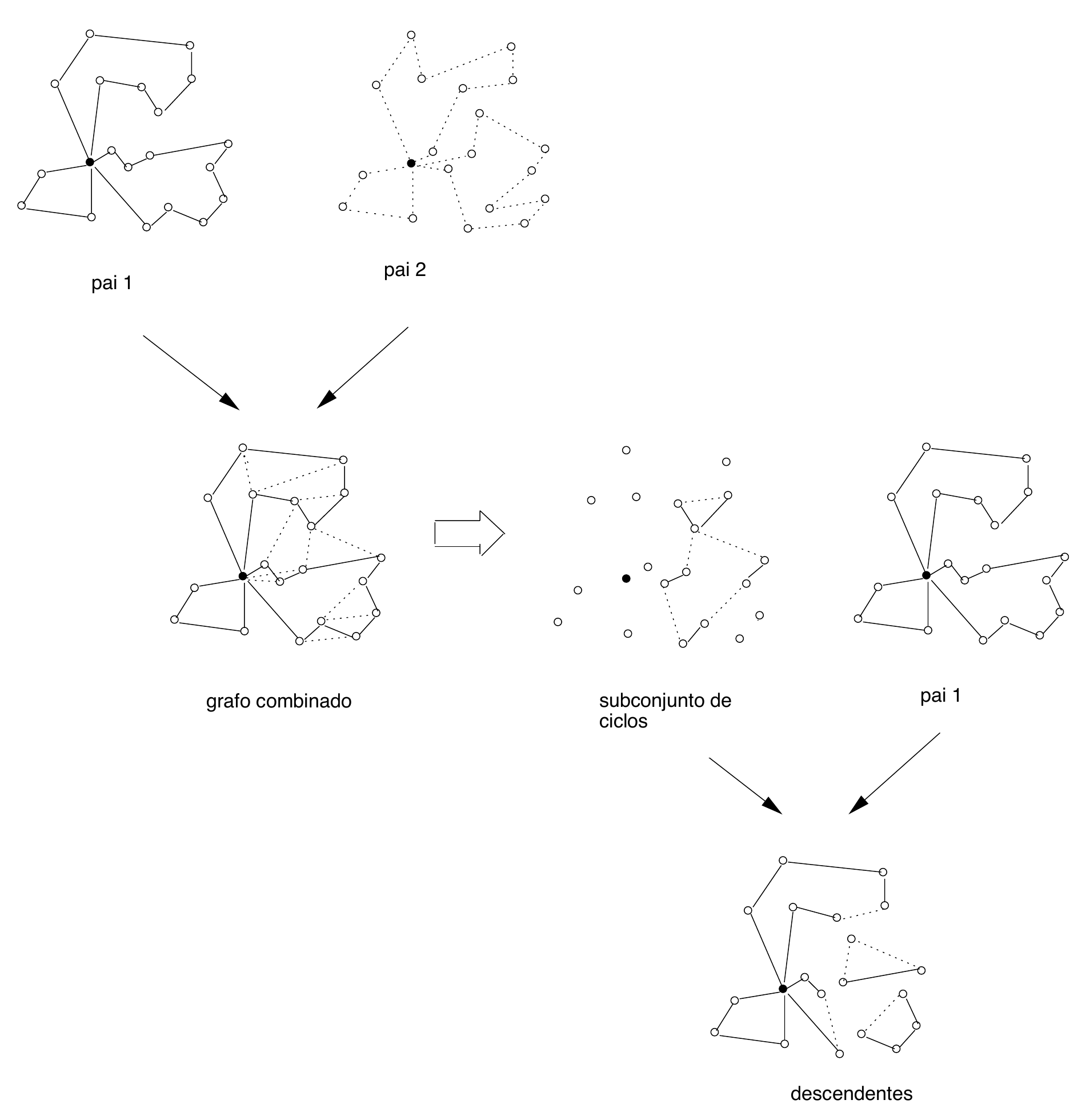}
}
  \caption{\emph{EAX crossover}. Imagem adaptada de~\cite{pereira2008bio}.}
  \label{fig:eax}
  \end{figure}

Como ilustrado na figura~\ref{fig:eax}, tal algoritmo começa gerando um grafo que tenha todas as arestas obtidas pela união de duas soluções pai. Subconjuntos de ciclos resultantes desta combinação são então particionados e um deles, assim como um dos pais, são selecionados para a etapa seguinte. O pai escolhido tem suas arestas presentes no subconjunto removidas enquanto as demais arestas do subconjunto lhe são adicionadas, criando uma solução intermediária. Caso o grafo descendente esteja desconexo, uma abordagem gulosa é utilizada para torná-lo conexo. Finalmente, violações de restrições são eliminadas e técnicas de troca de vértices ou $\lambda$-opt são aplicadas até que a solução seja viável.

%
%
%
%


\subsection{Demais Técnicas de Otimização}
\label{sec:outras}
Poderão ser combinadas, por exemplo, técnicas de otimização  heurística baseadas em busca local~\cite{ceschia2013local,mladenovic1997variable}, \emph{simulated annlealing} e busca tabu~\cite{osman1993metastrategy}. Tais técnicas procuram varrer o domínio de soluções em direções 
de melhora da função objetivo. Cada permutação de $\sol(C,k)$ é uma solução possível e soluções próximas são consideras vizinhas.
 Tais técnicas varrem parte do domínio, saltando de solução em solução, buscando melhorar a função objetivo.
 
A técnica de \emph{branch-and-bound} também pode ser usada para o VRP~\cite{diego2015phd}. Tal técnica pertence a classe de algoritmos exatos. Ela varre, de maneira organizada, o domínio de solução, onde grandes conjuntos são removidos no processo de poda.
Apesar de ter complexidade exponencial para encontrar a solução ótima, o  \emph{branch-and-bound}  possui um \emph{gap} entre a melhor solução encontrada e um limitante inferior para o ótimo. O \emph{gap}  pode ser útil mesmo que o    \emph{branch-and-bound} não execute até o fim. Algoritmos genéticos e a maioria das heurísticas descritas neste projeto não fornecem nenhuma informação de quão longe a solução se encontra do ótimo.

Além de \emph{branch-and-bound}, iremos buscar etabelecer limitantes inferiores para o problema por meio de relaxações. Por exemplo, o TSP é, em geral, uma relaxação para o VRP. Iremos buscar outras relaxações que nos ajudem a definir quão longe estamos da solução ótima.

\subsection{Objetivos}
\label{sec:objetivos}

(ajustar esta parte, era para qualificação (futuro) e agora passou a ser dissertação (o que foi feito))

Esta pesquisa visou estudar diversas técnicas de otimização e projetar algoritmos para o
\vrp. Foi trabalhada a variante do problema onde o comprimento das rotas é limitada por uma constante e foco foi dado á minimização do número de veículos, seguido pelo comprimento total de rotas.



{Recentemente Meira et al.~\cite{2016arXiv161005402Z} propuseram uma ferramenta para geração de instâncias 
do VRP. Tal ferramenta foi utilizada para modelar um problema real: a entrega de correspondências por carteiros
a pé na cidade de Artur Nogueira. Neste problema, são considerados em torno de 25 carteiros e milhares de 
correspondências a serem entregues por dia. As distâncias são medidas em tempo e existe um limite para o 
comprimento da rota. Por exemplo, uma rota deve ter comprimento máximo de 8h, definida a partir da jornada 
de trabalho de um carteiro.

Em tal variante, o carteiro não possui capacidade, uma vez que existe um veículo de apoio que reabastece os carteiros
quando necessário. O problema proposto é multi-objetivo. Deseja-se minimizar o comprimento
médio das rotas. Deseja-se minimizar o número de carteiros e, finalmente, deseja-se minimizar a variação dos comprimentos
das rotas. }

Este trabalho deu continuidade às pesquisas contidas em~\cite{2016arXiv161005402Z}. Foram estudados e propostos algoritmos  para obtenção de bons limitantes superiores bem como técnicas combinatórias para geração de 
limitantes inferiores. 

Resultados obtidos foram avaliados frente resultados disponíveis na literatura.
Além do benchmark dos correios, um outro conjunto de instâncias foi utilizado para fins de comparação.


\section{Síntese da Bibliografia Fundamental}
O Problema do Caixeiro Viajante (\tsp) é um dos mais estudados da otimização combinatória~\cite{Applegate99chainedlin-kernighan,applegate2006traveling,cook2012pursuit,lawler1985traveling}.
Uma generalização largamente estudada do \tsp~é o  Problema de Roteamento de Veículos, 
também conhecido como \emph{Vehicle Routing Problem (\vrp)}~\cite{toth2001vehicle}. Neste problema, deseja-se
encontrar $k$ ciclos que cubram todos os pontos de entrega (clientes) e que possuam  um vértice especial em comum (depósito).
O VRP possui  requisitos adicionais de acordo com a situação. 

Uma das primeiras aparições do VRP foi em 1959~\cite{dantzig1959truck} sob o nome de \emph{Truck Dispaching Problem}, uma generalização do Problema do Caixeiro Viajante.
 O nome VRP aparece em 1976 no tabalho de Christophides~\cite{christofides1976vehicle}. Christophides define VRP como um nome genérico dado a uma classe de problemas que envolvem visitar ``clientes'' com veículos.
 
 Situações distintas levam a diversas variantes do problema. Assumindo-se capacidade no veículo, temos o \textit{Capacitated-VRP (CVRP)}~\cite{fukasawa2006robust}. Definindo-se janelas temporais de entrega, temos o  \textit{VRP with Time Windows (VRPTW)}~\cite{kallehauge2005vehicle}. Havendo mais de um depósito, temos o \textit{Multi-Depot VRP (MDVRP)}~\cite{renaud1996tabu}. Outras variantes são facilmente encontradas na literatura.

O VRP modela diversas situações reais, como o roteamento de veículos para atender chamadas ou ocorrências,
definição de rotas para carteiros, definição de rotas para coleta de lixo, definição de itinerários de ônibus fretados e vans, entre muitas outras.
Ele é um problema da classe NP-Difícil \cite{CrescenziK00}, o que implica não existirem algoritmos eficientes para resolvê-lo na exatidão a menos que $P=NP$.


Solomon~\cite{solomon1987algorithms}, em 1987, criou um benchmark para VRP com janelas temporais de entrega. São 6 conjuntos de problemas num total de 56 instâncias, nas quais o número de clientes é sempre 100.  Os veículos possuem capacidade  de entregas e os clientes possuem demanda ou peso. O número de veículos não é fixo, mas deriva do fato de haver um limite de capacidade. Neste sentido, o problema pode ser considerado multi-objetivo. Deseja-se minimizar o percurso e o número de veículos.

%
%

Em 2014, Uchoa et. al. criou uma bilblioteca chamada CVRPLib~\cite{CVRPLIB}. Nesta biblioteca, eles consolidaram as  instâncias para o CVRP dos trabalhos~\cite{augerat1995computational,christofides1969algorithm,christofides1979vehicle,fisher1994optimal,golden1998impact,li2005very}. Além disso, Uchoa et. al~\cite{uchoanew} criaram   novas instâncias com número de clientes variando de 100 a 1000.


%
%
%

Kim et. al~\cite{Kim2015} fizeram  um levantamento 
do estado da arte sobre City VRP. Nesta variante existem restrições como, por exemplo,
 tráfego, poluição e  vagas para estacionar. O trabalho categoriza  mais de cem artigos sobre o VRP, levando em conta a logística da cadeia de produção em centros urbanos.

%

A tese de doutorado de Diego Aragão~\cite{diego2015phd}, defendida no departamento de informática da PUC-RJ em 2014, propõem algoritmos exatos para o CVRP. 
Nela, os ótimos das instâncias \texttt{M-n200-k16} e \texttt{M-n200-k17} foram encontrados pela primeira vez, por
meio de técnicas combinadas de geração de cortes e geração de colunas (\emph{Branch-Cut-and-Price}). 
Tais instâncias foram propostas por Christofides, Mingozzi e Toth em 1979~\cite{CVRPLIB,diego2015phd}.

O VRP também pode ser combinado com problemas de empacotamento.
Em 2016, A tese de doutorado de Pedro Hokama \cite{hokama2016}, defendida no IC-Unicamp, 
propôs algoritmos para o VRP combinado com empacotamento de paralelepípedos em \emph{containers}.

Diversos outros trabalhos tratam de roteamento combinado com empacotamento~\cite{Gendreau2008,Zachariadis2009729,dirk2016}.
Gendreau \cite{Gendreau2008} usa busca tabu para resolver o VRP combinado com empacotamento 2D.
Zachariadis \cite{Zachariadis2009729} utiliza meta-heurísticas como busca tabu associada a busca local para problemas de roteamento combinado com  empacotamento 2D.
%
Dirk \cite{dirk2016} cria algoritmos híbridos para solucionar roteamento combinado com empacotamento 3D.
%


Em 2008 foi lançado o livro \emph{Bio-inspired algorithms for the vehicle routing problem}~\cite{pereira2008bio}. Tal livro 
apresenta uma visão geral dos algoritmos bio-inspirados aplicados ao VRP.

\section{Metodologia e Implementação}


%
%

Considere uma instância $I$. Seja o número de entregas $n$. O conjunto de entregas pode ser representado por 
$Del=(1,\ldots,n)$.

\subsection{Algoritmo Genético}

Para este trabalho foi implementado um algoritmo genético e dois algoritmos incrementais de troca de arestas. O AG possui as etapas clássicas da metaheurísta: criação da população inicial, seleção dos indivíduos que serão cruzados, \emph{crossover}, mutação e seleção da nova população.

Após execução de testes preliminares,
foram definidos os seguinte parâmetros para o AG:

\bigskip
\centerline{
\begin{tabular}{ll}\hline
TAMANHO\_POPULACAO & 50 \\
NUMERO\_GERACOES & 100\\
TAXA\_CROSSOVER & 0.95\\
TAXA\_MUTACAO & 0.10\\
TAMANHO\_SELECAO\_CANDIDATOS & 50\\
\hline
\end{tabular}
}
\bigskip

As soluções são sequências de inteiros, um inteiro para cada ponto de entrega. Cada solução $Sol$ é uma permutação do conjunto $Del$,
ou seja $Sol\in Per(Del)$.
Na representação escolhida, não há delimitação de rotas e o depósito não está incluído. As rotas são delimitadas a posteriori por um algoritmo guloso.

\textbf{Demarcação de rotas:} A criação de um cromossomo $C$ é feita da seguinte maneira.
Os elementos de $Sol$ são inseridos sequencialmente no final de uma rota até que o valor limite para cada rota estabelecido pela instância não permita novas inserções. Quando isto ocorre, uma nova rota é criada e o processo se repete até todos os elementos serem inseridos. Feito isso, a solução passa a ser representada por sequências de inteiros intercaladas por zeros, que marcam o início/fim de cada rota. Note que o inicio e fim de cada rota ocorre no depósito.

A população inicial é criada através de permutações aleatórias de $Del$ e a seleção dos candidatos ao \emph{crossover} é feita por torneio. O torneio é feito da seguinte maneira. São selecionados dois cromossomos aleatoriamente e o melhor entra na população de candidatos ao \emph{crossover}. Isso se repete até que a população de candidatos atinja o valor pré-definido. É permitido repetição no grupo de candidatos.

Foram implementados dois operadores de \emph{crossover}, \emph{Best Cost Route Crossover} (BCR)~\cite{OmbukiBerman2006MultiObjectiveGA} e \emph{Order Based Crossover} (OX)~\cite{Oliver:1987:SPC:42512.42542}, que serão descritos a seguir.

\begin{figure}
\includegraphics[scale=.6]{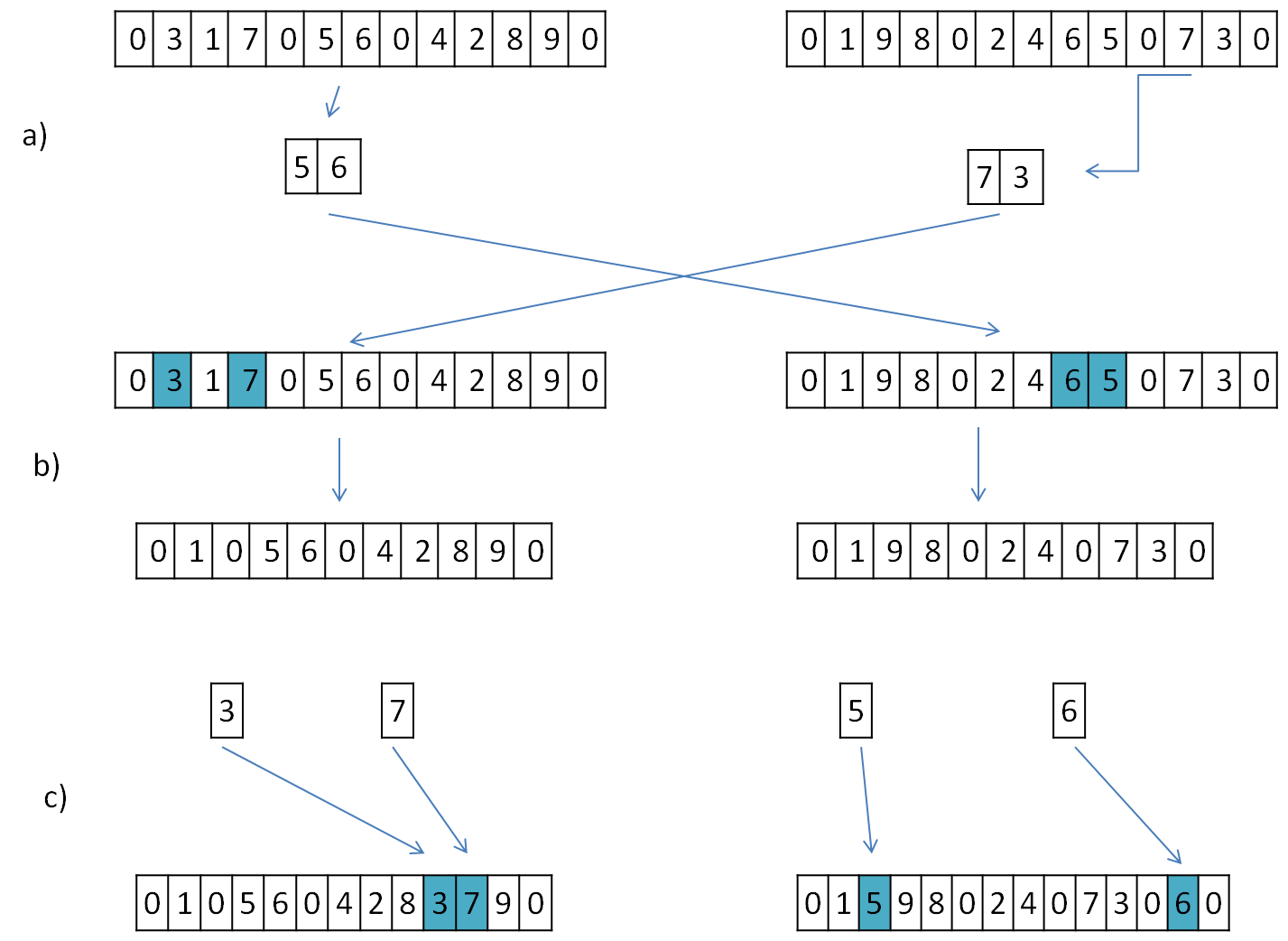}
\caption{Funcionamento do BCR~\emph{crossover} padrão. Adaptado de~\cite{OmbukiBerman2006MultiObjectiveGA}.}
\label{fig:bcrc}
\end{figure}

O BCR crossover foi desenvolvido originalmente para problemas multiobjetivo de VRP com janelas temporais e foi aplicado neste trabalho sem nenhuma adaptação. Para cada par de soluções pai, os seguintes passos são tomados. 
\begin{enumerate}
\item Uma rota é selecionada aleatóriamente de cada pai (Figura~\ref{fig:bcrc} a);
\item Os vértices pertencentes à rota selecionada de um pai são removidos do outro pai e vice-versa (Figura~\ref{fig:bcrc} b);
\item Os vértices removidos de cada pai são reinseridos, um por vez, de maneira independente, na melhor posição possível considerando todas as rotas. Caso necessário, o vértice é inserido em uma nova rota (Figura~\ref{fig:bcrc} c);.
\end{enumerate}

Como ilustrado em \ref{fig:bcrc}, no primeiro passo uma rota de cada pai é selecionada. No segundo passo é realizada a remoção de vértices e em seguida, no terceiro passo, os vértices são reinseridos.

A implementação utilizada difere levemente do BCR padrão, pois após a remoção dos vértices foi feita uma nova demarcação de rotas.

\begin{figure}
\includegraphics[scale=.6]{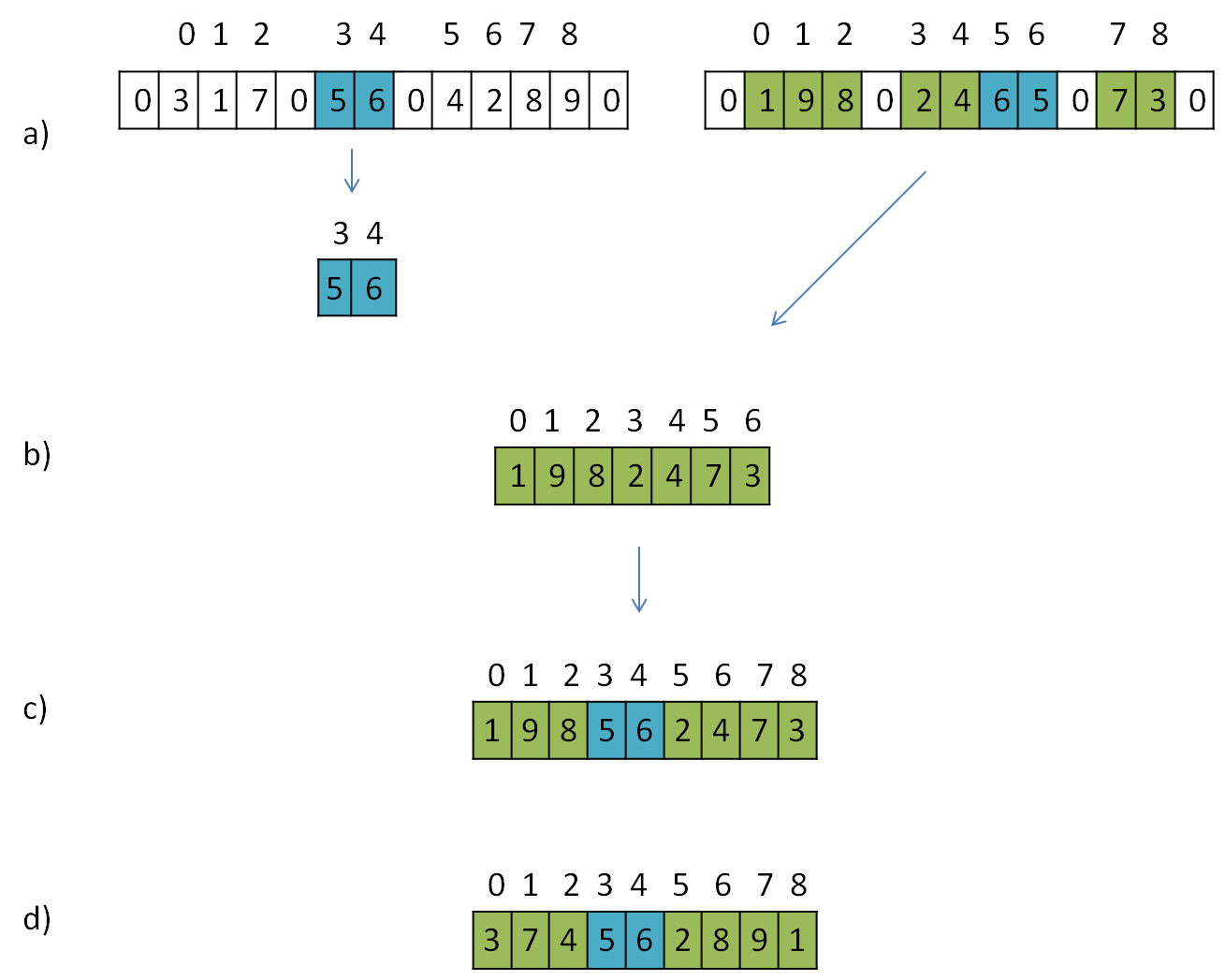}
\caption{Funcionamento do OX~\emph{crossover}.}
\label{fig:ox}
\end{figure}

O OX crossover foi desenvolvido originalmente para TSP e foi aplicado neste trabalho com uma adaptação para lidar com rotas ao invés de cortes. Para cada par de soluções pai, os seguintes passos são tomados. 
\begin{enumerate}
\item Uma rota é selecionada aleatóriamente de um dos pais, pai A (esquerda) (Figura \ref{fig:ox} a);
\item Um filho sem delimitação de rotas é criado inicialmente apenas com os vértices do pai B (direita), em sua ordem original, excluindo os vértices presentes na rota selecionada do pai A (esquerda) (Figura \ref{fig:ox} b);
\item Os vértices da rota selecionada do pai A (esquerda) são inseridos no filho a partir da mesma posição em que estariam no pai A (esquerda) (Figura \ref{fig:ox} c);
\item Um segundo filho é criado da mesma maneira, mas com os vértices do pai B em ordem inversa (Figura \ref{fig:ox} d).
\end{enumerate}

Para a etapa de mutação, três vértices aleatórios são selecionados da sequência de inteiros sem demarcação de rotas. Estes vértices são trocados de posição entre si e é feita uma nova demarcação de rotas. A melhor combinação é usada no cromossomo. Note que a mutação pode piorar o custo da rota.

Finalizando o AG, após a etapa de mutação, a população é atualizada com os $x$ melhores indivíduos sobrevivendo para a próxima geração, sendo $x$ o tamanho definido previamente.

A função objetivo usada no calculo de \emph{fitness} para seleção dos melhores indivíduos funciona da seguinte forma. Primeiro é considerado o número de veículos da solução. Em caso de empate, é considerado o comprimento da menor rota. critérios de parada: 100 gerações ou 20 gerações sem melhoria.

Além do AG, dois algoritmos incrementais de troca de arestas foram utilizados. O primeiro algoritmo foi o 2-opt. Trata-se de um processo determinístico que testa, para cada rota de uma solução, todas as trocas de pares de arestas possíveis e realiza a melhor delas até que o custo da rota não possa mais melhorar.

O segundo algoritmo é uma variação de \emph{String Exchange}, o qual executa trocas de sequências de vértices entre rotas de uma solução da seguinte forma (\emph{String Exchange'}).

\begin{figure}
\includegraphics[scale=.6]{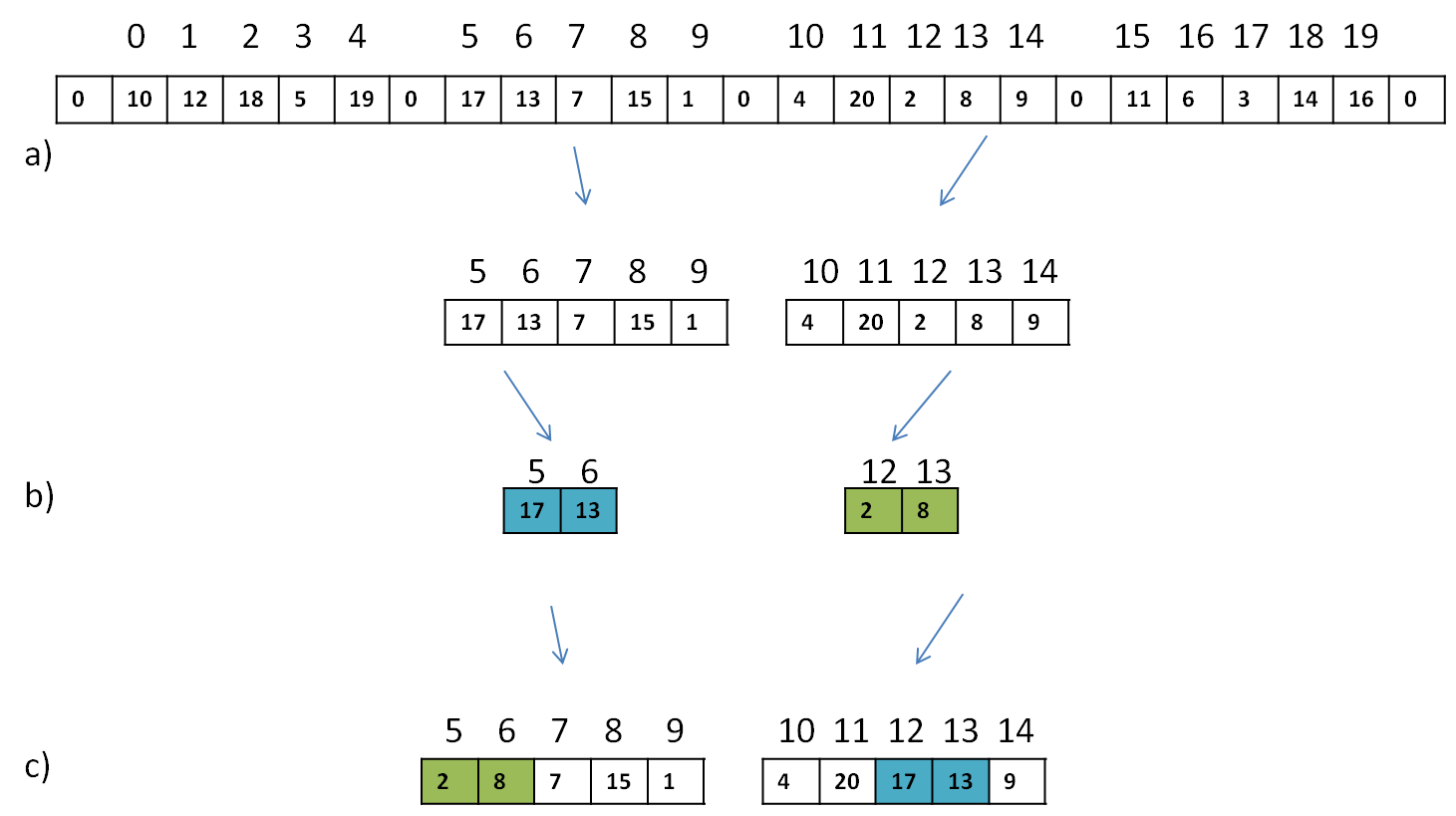}
\caption{Funcionamento do~\emph{String Exchange}.}
\label{fig:edgeexch}
\end{figure}

\begin{enumerate}
\item Duas rotas são selecionadas aleatóriamente da solução (Figura~\ref{fig:edgeexch} a);
\item Um seguimento de tamanho pré definido $n$ é extraído de uma posição aleatória de cada rota (Figura~\ref{fig:edgeexch} b);
\item O seguimento extraído de uma rota é trocado pelo seguimento da outra rota (Figura~\ref{fig:edgeexch} c);
\item Caso haja melhora da solução, a troca é mantida e os passos anteriores se repetem $x$ vezes;
\item Após $x$ iterações, todo o processo recomeça com $n\gets n-1$, até que $n=0$.
\end{enumerate}

Os valores iniciais de $n$ e $x$ para este algoritmo foram 3 e 30000, respectivamente.

Na parte experimental foram feitas duas variantes de AG. Na primeira, a busca local 2-opt foi aplicada à população resultante ao final de cada geração do AG. Na segunda, não foi aplicado o 2-opt.
Em todos os experimentos, o~\emph{String Exchange} é aplicado ao melhor indivídiduo produzido pelo AG.

\subsection{Implementação \emph{Multi-thread}}

%
%
%
%

\subsection{Instâncias Utilizadas}

Quatro conjuntos de instâncias foram utilizados neste trabalho.
Os conjuntos {RealWorldVRP}, {ManhattanPilot} e o {Example} obtidos em~\cite{zenipostvrp,2016arXiv161005402Z} para o PostVRP e o conjunto X \cite{uchoanew} para CVRP.

O conjunto para PostVRP foi desenvolvido a partir de um estudo de caso envolvendo os correios da cidade de Artur Nogueira onde em torno de $25$ carteiros realizam até $30.000$ entregas por dia. Neste problema os carteiros a pé são os veículos e as ruas são arestas. Cada instância modela um dia de trabalho dos carteiros cujo expediente não pode ultrapassar 6-8 horas. Isso limita o comprimento das rotas, que é medido em unidades de tempo.

As 78 instâncias do conjunto RealWorldVRP estão divididas em quatro grupos. O grupo \emph{Toy} possui instâncias menores, de até $5.000$ pontos de entrega, para validação de algoritmos. O \emph{Normal} possui casos mais realistas, de $10.000$ a $14.000$ pontos. Por fim, os grupos \emph{OnStrike} e \emph{Christmas} representam situações mais atípicas.

 Adicionalmente, a ferramenta usada para gerar as instâncias dos correios também oferece os conjuntos \emph{example} e \emph{ManhattanPilot}, de até 10.000 pontos.
 
 O conjunto X foi proposto em 2017 para complementar os já consolidados conjuntos do CVRP clássico. Estas instâncias, que podem ter até 1.000 pontos de entrega, visam oferecer uma diversidade maior de problemas, promover o desenvolvimento de métodos mais flexíveis e permitir sua extensão para uso em outras variantes de VRP. Neste trabalho, tal conjunto foi usado principalmente por contar com ótimos conhecidos, proporcionando uma melhor avaliação dos métodos propostos.

\section{Resultados}

Todos os experimentos foram executados em um \emph{cluster} IBM com 2.6-3.4GHz e 768GB de memória (processo  FAPESP projeto 2010/50646-6). Foram utilizadas 24 threads e 32GB de RAM para cada processo.

\subsection{Testes Preliminares}

 Testes preliminares foram realizados no cluster para a definição manual de 
 parâmetros do AG.
 O primeiro teste consistiu em executar algumas instâncias do conjunto \texttt{example} variando o número de \emph{threads} ideal para uma execução rápida.
 O tamanho da população e a taxa de \emph{crossover} foram mantidos em $50$ e $0.95$, respectivamente. 

%
%


O acréscimo de \emph{threads} mostrou uma redução na duração total do AG até 24 \emph{threads}. 
A partir deste experimento, foi definido o número de threads como 24.

O segundo teste teve como objetivo definir o melhor tamanho de população de modo que a quantidade de indivíduos fosse grande o bastante para proporcionar melhores, mais variadas soluções, mas ainda pequena o suficiente para não inflar demais o tempo de execução do AG. 

Utilizando a instância \texttt{RealWorldPostToy-200-0} de tamanho 200 do conjunto \emph{RealWorld}, quatro tamanhos de população foram testados: 25, 50, 75 e 100.
A instância foi executada 40 vezes para cada tamanho de população, variando o operador de \emph{crossover} e o uso de busca local. Os detalhes sobre o experimento estão na Seção \ref{sec:experimentos}.

Os resultados indicaram qualidade parecida para todos os tamanhos de população, porém os tempos de execução foram $7$, $51$, $104$ e $191$
minutos para populações de tamanho 25, 50, 75 e 100, respectivamente.
O tamanho de população escolhido para os experimentos foi de 50 por proporcionar um bom equilíbrio entre resultados e velocidade.

Já o valor da mutação foi testado para a mesma instância \texttt{RealWorldPostToy-200-0} e tamanho de população 50. Foram testados os valores 5\%, 10\% e 15\%. O melhor resultado foi obtido para taxa de mutação igual a 10\%. Tal valor foi utilizado nos experimentos seguintes.

\subsection{Experimentos}
\label{sec:experimentos}

O principal experimento desta pesquisa envolveu comparar o desempenho do AG usando dois operadores diferentes de \emph{crossover} e o uso ou não uso de busca local. Foram então realizadas quatro configurações de experimentos de AG: 
\begin{itemize}
\item [(a)] Utilizando o operador BCR-Crossover com algoritmo 2-opt para rotas individuais aplicado ao final de cada geração para toda a população (BCR LS Enable); 
\item [(b)] Utilizando o operador BCR-Crossover sem busca local 2-opt (BCR LS Disable);  
\item [(c)] Utilizando o operador OB-Crossover com algoritmo 2-opt para rotas individuais aplicado ao final de cada geração para toda a população (OX LS Enable). 
\item [(d)] Utilizando o operador OB-Crossover sem busca local 2-opt (OX LS Disable); 
\end{itemize}
Lembrando que para todos os casos, ao final do AG, é feita a busca local \emph{String Exchange}
na melhor solução encontrada.

Cada instância foi executada dez vezes em cada uma das quatro configurações, totalizando quarenta execuções por instância.



\subsection{Instâncias do PostVRP}

Nos experimentos descritos a seguir, utilizamos as instâncias descritas em~\cite{2016arXiv161005402Z}. 
Tais instâncias são uma versão multi-objetivo do VRP com comprimento máximo da rota limitado.
 Neste benchmark, o nome da instância é composta por três partes. Por exemplo, a instância \texttt{ManhattanPilot\_500\_2} pertence ao grupo ManhattanPilot, possui 500 pontos de entrega e possui identificador $ID=2$.

\begin{figure}[htb]
\begin{center}
\includegraphics[width=1\linewidth]{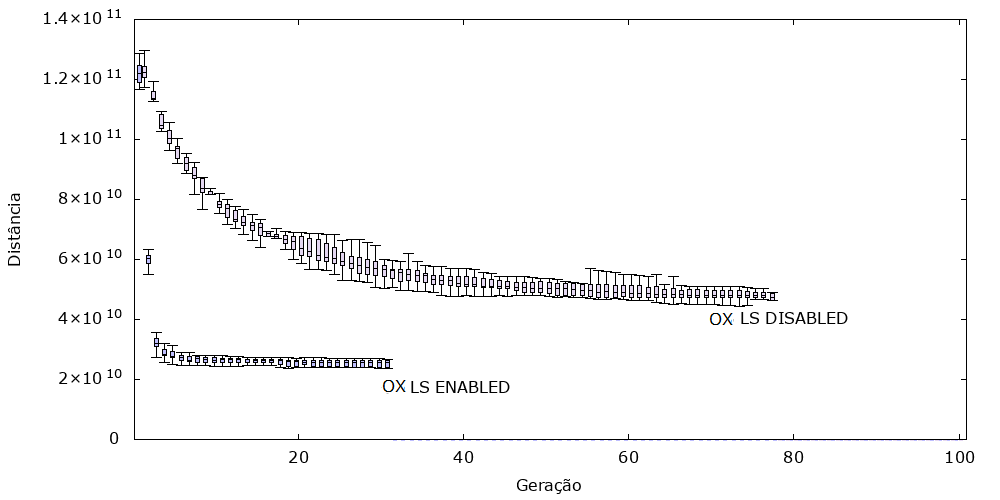}
\includegraphics[width=1\linewidth]{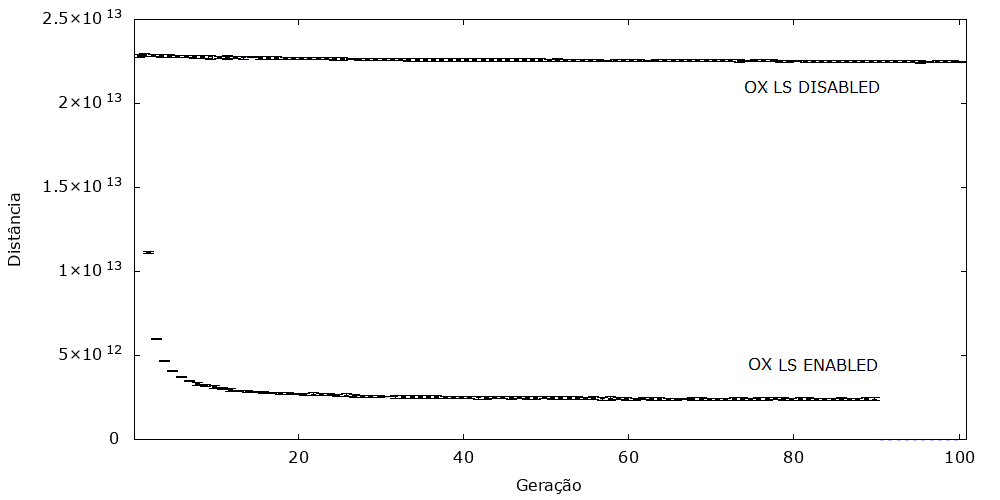}
\end{center}
\caption{Distância versus geração para as instâncias \texttt{RealWorldPostToy-100-1} (acima) e \texttt{example-10000-5} (abaixo), usando o operador OB com e sem busca local ao final de cada geração.}
\label{fig:lsacelera}
\end{figure}

\begin{figure}[htb]
\begin{center}
\includegraphics[width=1\linewidth]{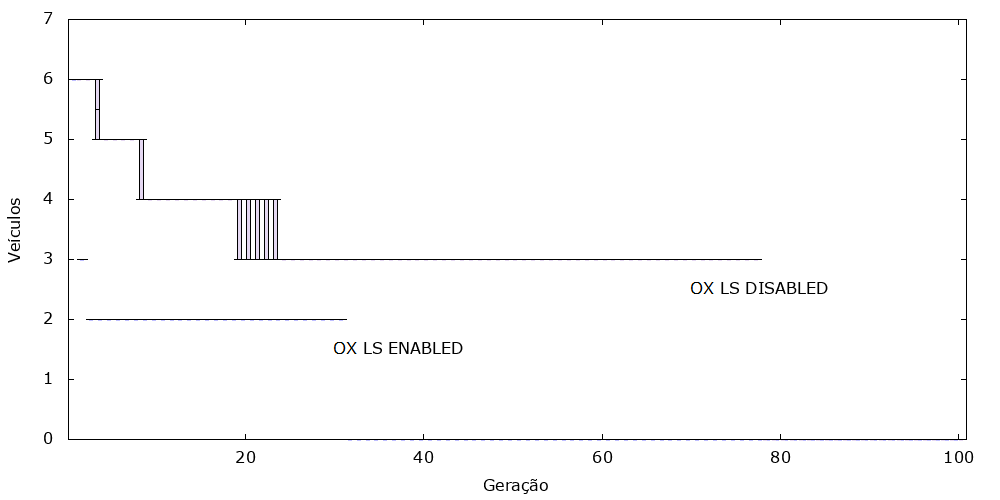}
\includegraphics[width=1\linewidth]{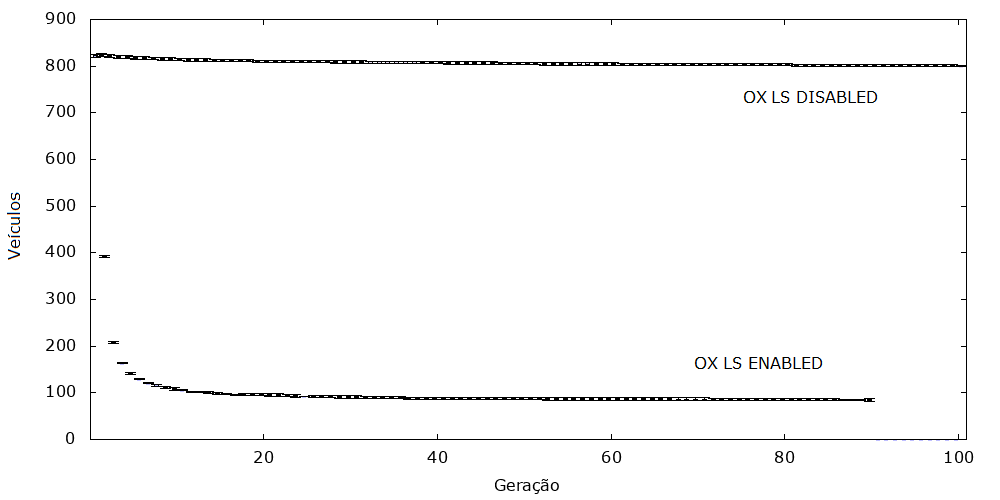}
\end{center}
\caption{Número de veículos versus geração para as instâncias \texttt{RealWorldPostToy-100-1} (acima) e \texttt{example-10000-5} (abaixo), usando o operador OB com e sem busca local ao final de cada geração.}
\label{fig:lsacelerak}
\end{figure}

Nas figuras~\ref{fig:lsacelera} e~\ref{fig:lsacelerak}  comparamos o operador OX LS Enable com OX LS Disable para uma instância de tamanho 100 e para outra de tamanho 10.000. 
A Figura~\ref{fig:lsacelera} contém a distância \emph{versus} geração e a Figura~\ref{fig:lsacelerak} contém o número de veículos \emph{versus} geração.
Em ambos os casos, o operador de busca local trouxe uma convergência mais acentuada, porém a melhora é mais significativa na instâcnia com 10.000 entregas.
 
\begin{figure}[htb]
\begin{center}
\includegraphics[width=1\linewidth]{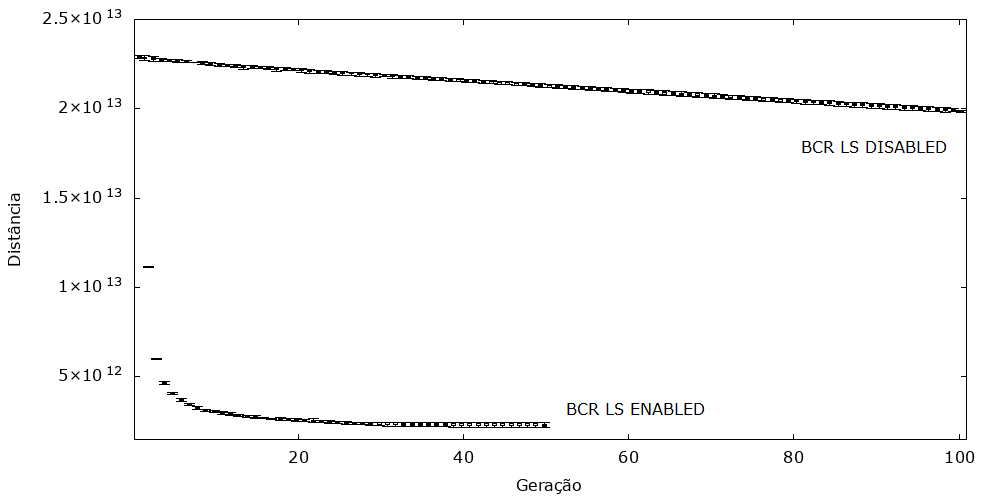}
\includegraphics[width=1\linewidth]{EX10000_oblsC2.png}
\end{center}
\caption{Distância versus geração para os operadores BCR (acima) e OX (abaixo), instância \texttt{example-10000-5}, com e sem busca local ao final de cada geração.}
\label{fig:bcreobls}
\end{figure}

\begin{figure}[htb]
\begin{center}
\includegraphics[width=1\linewidth]{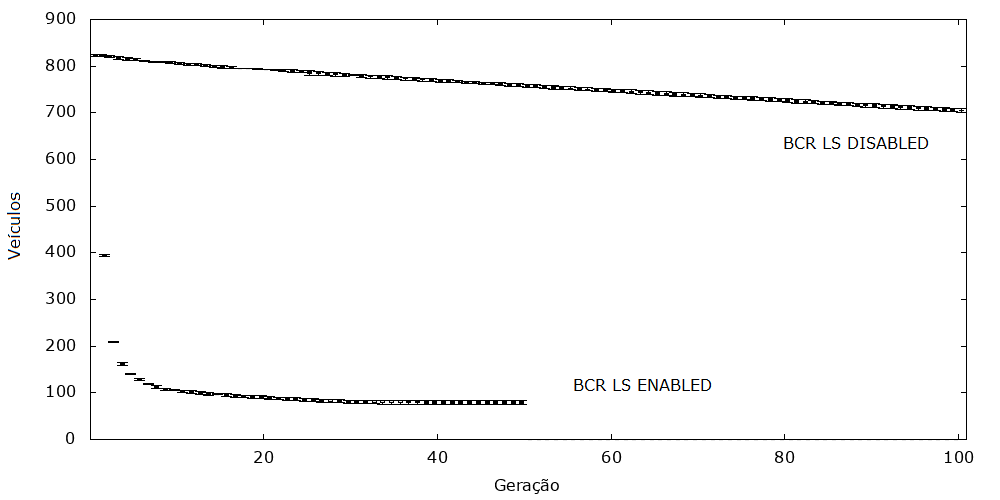}
\includegraphics[width=1\linewidth]{EX10000_oblsK2.png}
\end{center}
\caption{Número de veículos versus geração para os operadores BCR (acima) e OX (abaixo), instância \texttt{example-10000-5}, com e sem busca local ao final de cada geração.}
\label{fig:bcreoblsk}
\end{figure}

Nas figuras~\ref{fig:bcreobls} e~\ref{fig:bcreoblsk}  comparamos os operadores BCR e OX nas configurações LS Enable e LS Disable para uma instância de tamanho 10.000. 
A Figura~\ref{fig:bcreobls} contém a distância \emph{versus} geração e a Figura~\ref{fig:bcreoblsk} contém o número de veículos \emph{versus} geração.
Na Figura~\ref{fig:bcreoblsk}, o mínimo número de veículos para (OX LS Enable, OX LS Disable) foi 
$(82,799)$ respectivamente. Já para (BCR LS Enable, BCR LS Disable), o número mínimo de veículos foi
$(74,701)$. 
A melhora da convergencia para o LS Enable foi obtida para ambos os operadores. Também podemos observar que o operador BCR teve um desempenho  superior ao OX. 


\begin{figure}[htb]
\begin{center}
\includegraphics[width=1\linewidth]{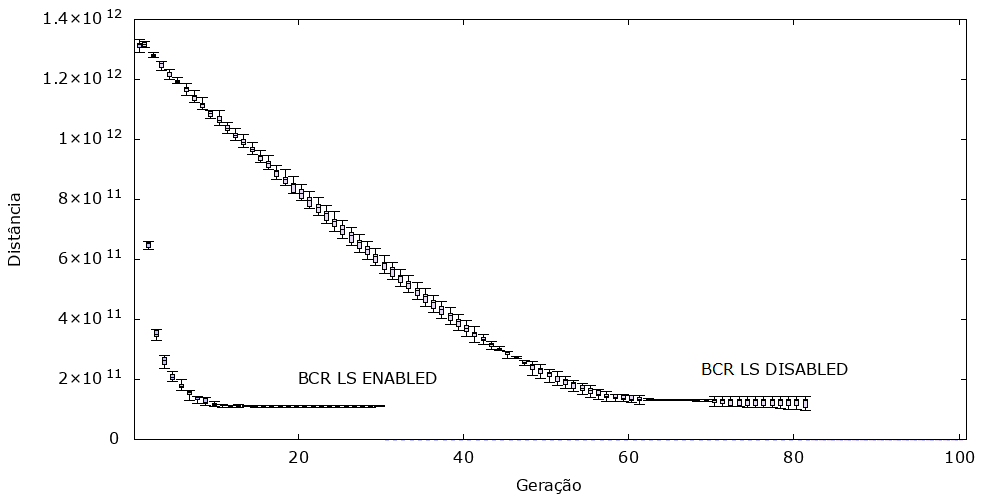}
\end{center}
\caption{Distância versus geração usando o operador BCR, instância \texttt{RealWorldPostToy-1000-2} com e sem busca local ao final de cada geração.}
\label{fig:bcrls}
\end{figure}

\begin{figure}[htb]
\begin{center}
\includegraphics[width=1\linewidth]{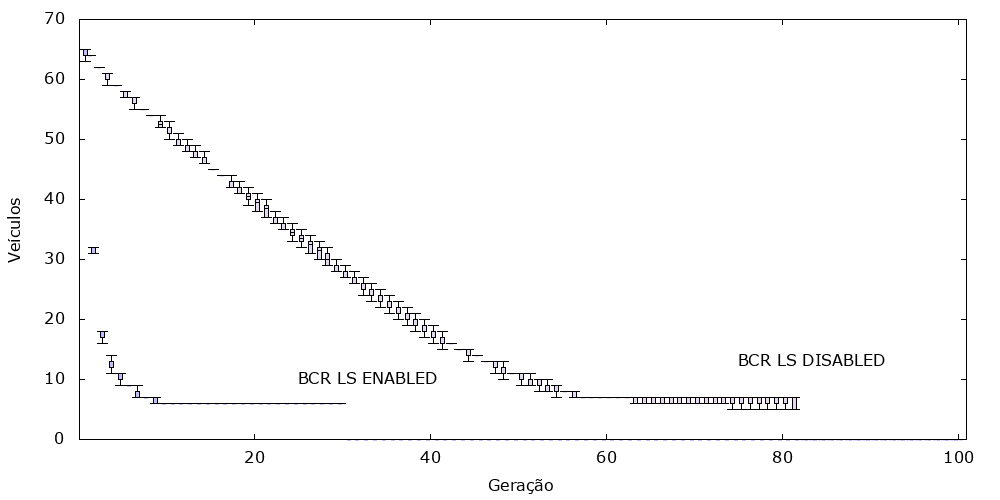}
\end{center}
\caption{Número de veículos versus geração usando o operador BCR, instância \texttt{RealWorldPostToy-1000-2} com e sem busca local ao final de cada geração.}
\label{fig:bcrlsk}
\end{figure}

Em todos os experimentos utilizamos dois critérios de parada: 100 gerações ou 20 gerações sem melhoria. Em algumas situações específicas, quando o tamanho da instância é pequeno, o algoritmo 
LS Disable
obteve resultado levememnte superior 
ao algoritmo LS Enable. 
Isso se deve ao comportamento estocástico do AG. Veja as figuras \ref{fig:bcrls} e \ref{fig:bcrlsk}.

\begin{figure}[htb]
\begin{center}
\includegraphics[width=1\linewidth]{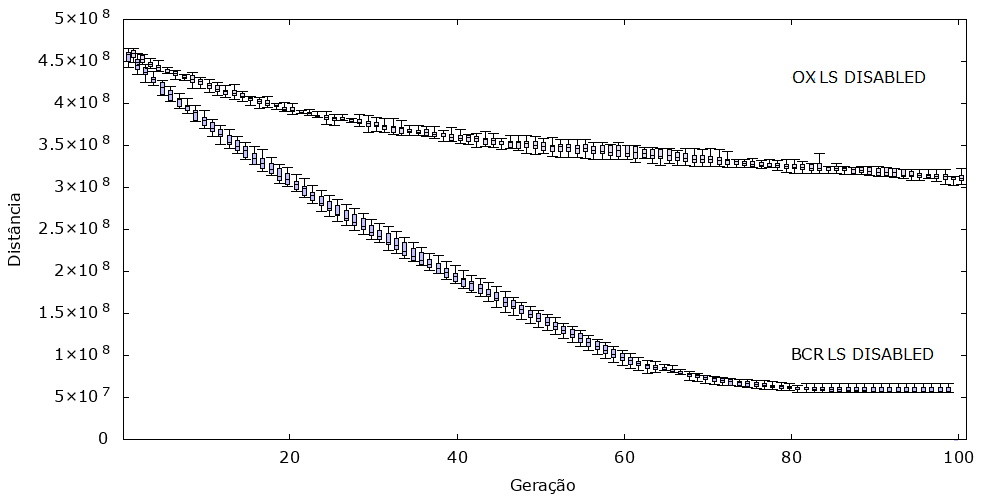}
\includegraphics[width=1\linewidth]{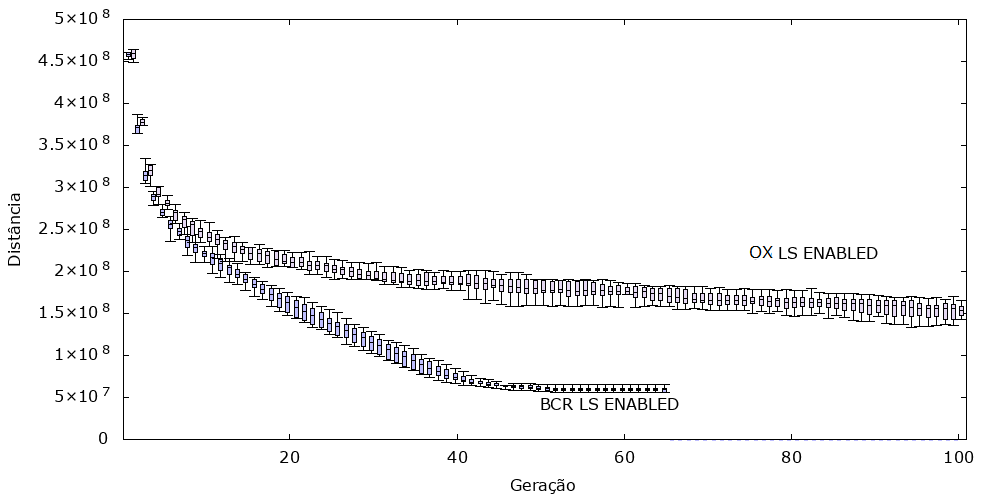}
\end{center}
\caption{Distância versus geração para os operadores BCR e OX, instância \texttt{ManhattanPilot-500-1}, sem busca local (acima) e com busca local (abaixo) ao final de cada geração.}
\label{fig:bcrob-bcrobls}
\end{figure}

\begin{figure}[htb]
\begin{center}
\includegraphics[width=1\linewidth]{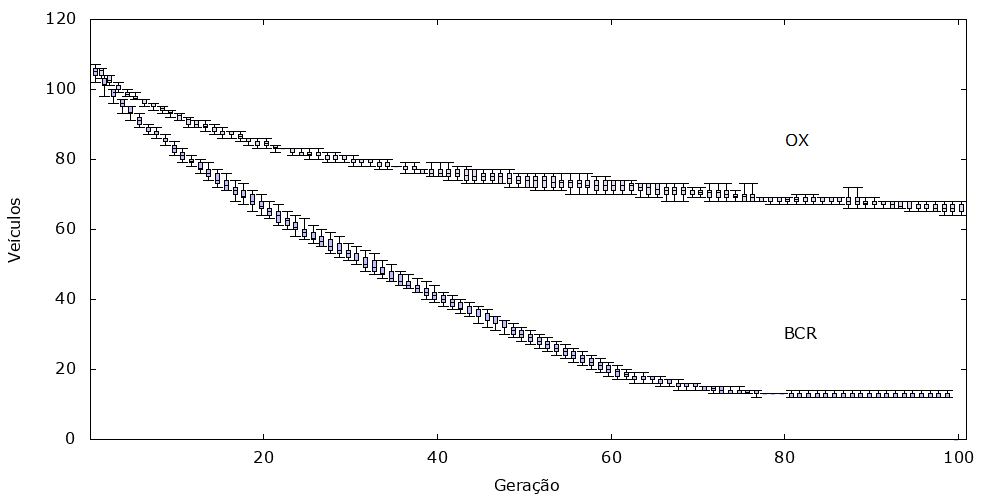}
\includegraphics[width=1\linewidth]{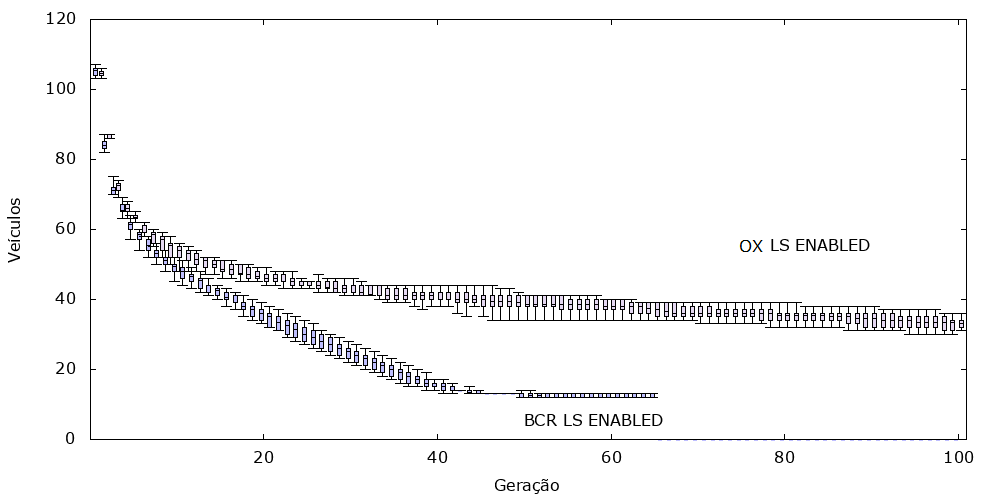}
\end{center}
\caption{Número de veículos versus geração para os operadores BCR e OX, instância \texttt{ManhattanPilot-500-1}, sem busca local (acima) e com busca local (abaixo) ao final de cada geração.}
\label{fig:bcrob-bcroblsk}
\end{figure}

\begin{figure}[htb]
\begin{center}
\includegraphics[width=1\linewidth]{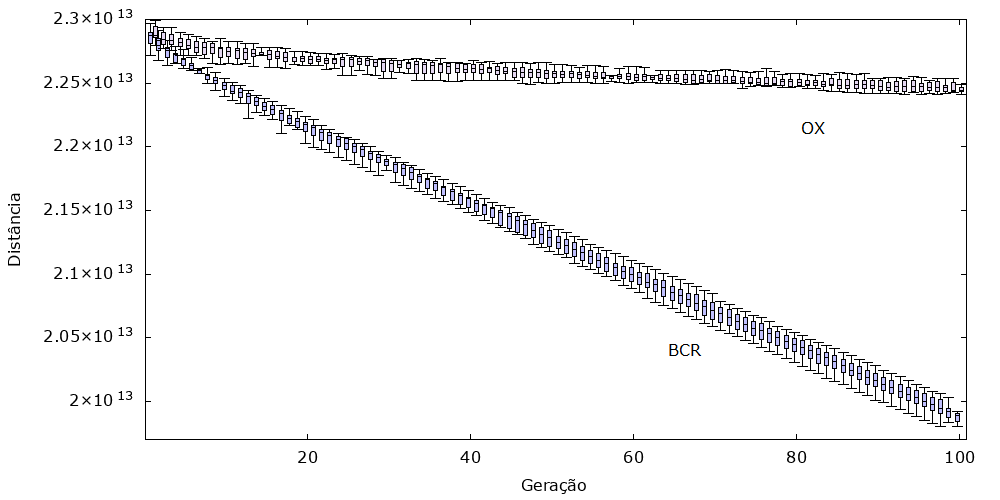}
\includegraphics[width=1\linewidth]{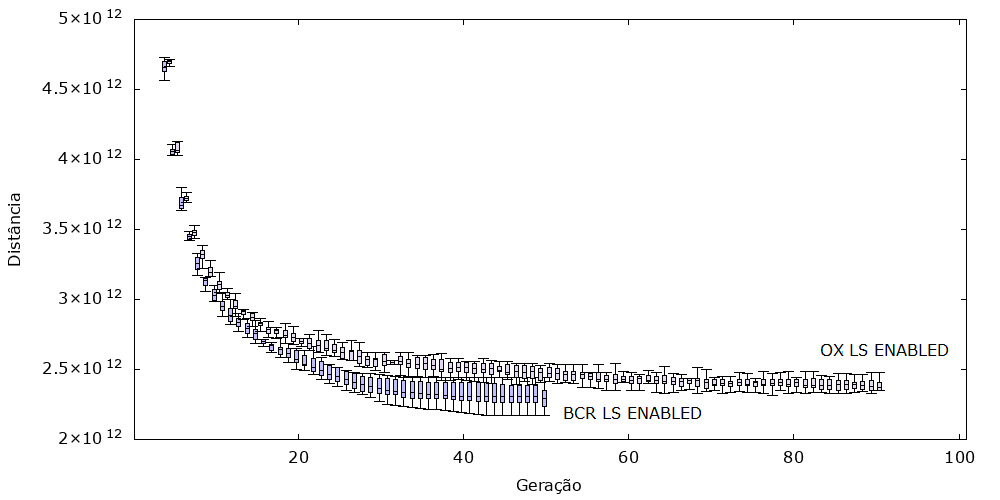}
\end{center}
\caption{Distância versus geração para os operadores BCR e OX, instância \texttt{example-10000-5}, sem busca local (acima) e com busca local (abaixo) ao final de cada geração.}
\label{fig:bcrob-bcrobls10000}
\end{figure}

\begin{figure}[htb]
\begin{center}
\includegraphics[width=1\linewidth]{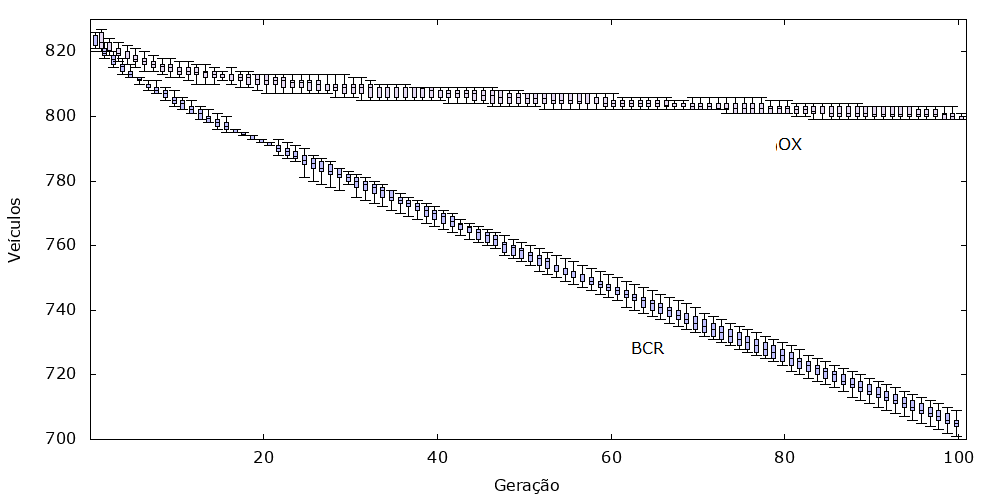}
\includegraphics[width=1\linewidth]{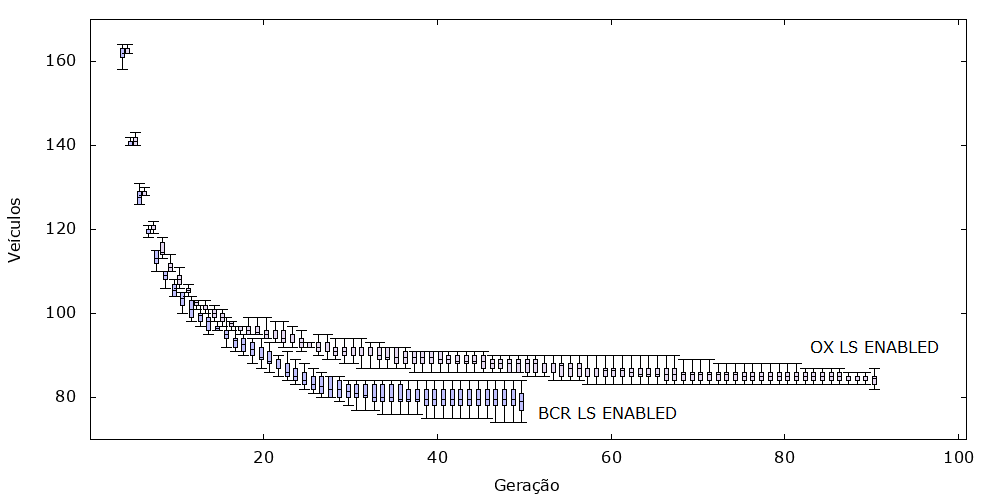}
\end{center}
\caption{Número de veículos versus geração para os operadores BCR e OX, instância \texttt{example-10000-5}, sem busca local (acima) e com busca local (abaixo) ao final de cada geração.}
\label{fig:bcrob-bcrobls10000k}
\end{figure}

 Comparando os operadores de \emph{crossover}, o operador BCR se mostrou superior ao OX na grande maioria dos casos.
Apesar disso, as figuras~\ref{fig:bcrob-bcrobls}
e~\ref{fig:bcrob-bcroblsk} mostram que com o auxílio do 2-opt aplicado a cada geração, o desempenho do operador OX consegue se aproximar mais do operador BCR. Para instâncias maiores, esta aproximação também parece crescer. Veja as figuras~\ref{fig:bcrob-bcrobls10000} e \ref{fig:bcrob-bcrobls10000k}. 

Por ser mais simples, o AG com operador OX roda em torno de quatro vezes mais rápido que o AG com operador BCR.
Assim, o operador OX continua sendo uma opção competitiva, especialmente se combinado à busca local.

De um total de 113 instâncias descritas em~\cite{2016arXiv161005402Z} selecionamos  as 62 menores.
A Tabela~\ref{mainTable} contém a  solução com mínimo $(k,comprimento)$ encontrada nas 40 execuções, 10 para cada configuração distinta.
Foi comparado em primeiro lugar o valor $k$. Para duas soluções com mínimo $k$, foi comparado então a soma dos comprimentos das rotas.
Além dos custos, a Tabela mostra a configuração ou configurações que obtiveram o melhor resultado para cada instância.
Houve um empate dos quatro algoritmos para todas as instâncias com até 10 pontos de entrega. Tais instâncias são muito pequenas.
Para as demais instâncias, com 11 ou mais pontos de entrega, houve sempre uma configuração vitoriosa.
Tal configuração foi: \textit{(a)} BCR LS Enable em 26 casos, \textit{(b)} BCR LS Disable em 13 casos e \textit{(c)} OB LS Enable em 4 casos.

As instâncias com 500 pontos de entrega gastaram em torno de 2 horas de execução. Instâncas com 5000 pontos de entrega gastaram, no máximo, 93 horas. Cada execução rodou com até 24 threads em paralelo. Observe que o tempo diz respeito a soma das 40 execuções e foi medido ignorando-se o paralelismo.

A instância \texttt{RealWorldPostNormal-10000-0}, não presente na Tabela~\ref{mainTable}, foi executada apenas 16 vezes, 10 para BCR LS Disabled e 6 para BCR LS Enabled, devido a seu alto tempo de execução, que somou 15.137 minutos para estas execuções. Nestas circunstâncias, a configuração BCR LS ENABLED obteve o melhor resultado com 32 veículos e 688541690067 de comprimento.
(colocar um gráfico).

\begin{table}[htb]
\caption{Melhores resultados obtidos por instância. Seja (BCR LS Enable, BCR LS Disable, OB LS Enable, OB LS Disable) igual a (a,b,c,d) respectivamente. Mostramos na tabela o melhor $(k,comprimento)$, e o tempo das 40 execuções somados. Também destacamos qual algoritmo encontrou a melhor solução.}
\label{mainTable}
\scriptsize
\centerline{
\hspace{-1cm}\begin{filecontents*}{resultspvrp2.csv}
Instancia,Melhor K,Melhor Dist,Tempo,Instancia2,Melhor K2,Melhor Dist2,Tempo2
exp-10-5$^{abcd}$,1,15095885629,0,ManP-5000-1$^c$,511,954191742,1199
exp-100-5$^a$,2,35236853041,6,ManP-5000-2$^c$,499,2271821549,2494
exp-1000-5$^a$,8,2.1147E+11,318,RWPToy-3-0$^{abcd}$,1,6023937832,0
exp-10000-5$^a$,71,1.87672E+12,12138,RWPToy-3-1$^{abcd}$,1,6826653003,0
ManP-3-0$^{abcd}$,1,3709581,0,RWPToy-3-2$^{abcd}$,1,4127370448,0
ManP-3-1$^{abcd}$,1,2485357,0,RWPToy-5-0$^{abcd}$,1,3742109767,0
ManP-3-2$^{abcd}$,1,3786186,0,RWPToy-5-1$^{abcd}$,1,3461359665,0
ManP-5-0$^{abcd}$,1,4686357,0,RWPToy-5-2$^{abcd}$,1,5188914899,0
ManP-5-1$^{abcd}$,1,2686480,0,RWPToy-10-0$^{abcd}$,1,8103109376,0
ManP-5-2$^{abcd}$,1,4134533,0,RWPToy-10-1$^{abcd}$,1,8727033996,0
ManP-10-0$^{abcd}$,2,6575903,0,RWPToy-10-2$^{abcd}$,1,6082601657,0
ManP-10-1$^{abcd}$,1,4497191,0,RWPToy-20-0$^b$,1,11229149347,1
ManP-10-2$^{abcd}$,1,4678544,0,RWPToy-20-1$^b$,1,11253026628,1
ManP-20-0$^a$,2,7202261,0,RWPToy-20-2$^b$,1,9357553819,1
ManP-20-1$^a$,2,6982255,0,RWPToy-50-0$^a$,1,17762849447,3
ManP-20-2$^b$,2,8728363,0,RWPToy-50-1$^a$,1,13197120948,3
ManP-50-0$^a$,3,12611093,2,RWPToy-50-2$^c$,1,15646200132,3
ManP-50-1$^a$,3,10813610,1,RWPToy-100-0$^a$,2,23439941986,7
ManP-50-2$^b$,3,11945041,1,RWPToy-100-1$^a$,1,21404038732,7
ManP-100-0$^b$,4,16962018,5,RWPToy-100-2$^a$,1,21565328275,7
ManP-100-1$^b$,4,16948889,5,RWPToy-200-0$^a$,2,33590426863,29
ManP-100-2$^a$,4,15728046,5,RWPToy-200-1$^a$,2,33655199969,32
ManP-200-0$^b$,6,26546522,15,RWPToy-200-2$^a$,2,32211406237,33
ManP-200-1$^a$,6,26249777,15,RWPToy-500-0$^a$,3,61231448742,150
ManP-200-2$^b$,6,26257130,15,RWPToy-500-1$^b$,4,62710650345,153
ManP-500-0$^b$,11,49087528,71,RWPToy-500-2$^a$,3,60970050148,154
ManP-500-1$^b$,12,52109723,73,RWPToy-1000-0$^a$,5,99477299207,598
ManP-500-2$^a$,11,48930341,76,RWPToy-1000-1$^a$,5,94550435967,659
ManP-1000-0$^a$,18,87619991,181,RWPToy-1000-2$^b$,5,89635383711,644
ManP-1000-1$^a$,17,79051618,296,RWPToy-5000-1$^a$,17,367078707785,5296
ManP-1000-2$^a$,20,88791674,336,RWPToy-5000-2$^a$,18,387841691495,4938
ManP-5000-0$^c$,525,2381307163,1372,,,,


\end{filecontents*}
\csvreader[separator=comma, tabular=|lrrr|lrrr|, table head=\hline Inst\^{}\{alg.\} &  $\min k$ & $\min$ compr. & t (min.) &Inst\^{}\{alg.\}  &  $\min k$ & $\min$ compr. & t (min.)\\\hline, table foot=\hline]{resultspvrp2.csv}{Instancia=\inst,Melhor K=\avgk, Melhor Dist=\dist, Tempo=\time,
Instancia2=\instb,Melhor K2=\avgkb, Melhor Dist2=\distb, Tempo2=\timeb
}{\inst & \avgk & \dist & \time & \instb & \avgkb & \distb & \timeb}
}
\normalsize
\end{table}

\subsection{Instâncias do Conjunto X}

Todas as instâncias utilizadas até o momento não possuem limitantes conhecidos. Sendo assim, este trabalho também realizou o mesmo experimento utilizando as instâncias do ``Conjunto X''~\cite{uchoanew}, as quais possuem ótimo conhecido. As instâncias deste conjunto são do CVRP.

Neste benchmark, o nome da instância possui o número de entregas e o número de veículos. Por exemplo, a instância \texttt{X-n101-k25} possui 101 pontos de entregas e 25 veículos.

A Tabela \ref{xtable} mostra a configuração que obteve o melhor resultado de $k$ para cada instância e os valores ótimos. Para este conjunto as configurações que obtiveram o melhor desempenho foram: \textit{(a)} BCR LS ENABLED em 57 casos, \textit{(c)} OX LS ENABLED em 38 casos e \textit{(b)} BCR LS DISABLED em 5 casos.

\begin{table}[htb]
\caption{Execução do AG nas instâncias do Conjunto X. Seja (BCR LS Enable, BCR LS Disable, OB LS Enable, OB LS Disable) igual a (a,b,c,d) respectivamente. Mostramos na tabela os valores de $k$ obtidos e os valores ótimos~\cite{uchoanew}. Também destacamos qual algoritmo encontrou a melhor solução.}
\label{xtable}
\footnotesize
\centerline{
\begin{filecontents*}{x2.csv}
ID,Instancia,Melhor K,K Otimo,n,ID2,Instancia2,Melhor K2,K Otimo2,n2,ID3,Instancia3,Melhor K3,K Otimo3,n3
X-n101-k25-BCR-Opt-4,X-n101-k25$^a$,26,25,1.04,X-n261-k13-OBC-Opt-4,X-n261-k13$^c$,13,13,1,X-n502-k39-BCR-Opt-9,X-n502-k39$^a$,42,39,1.076923077
X-n106-k14-OBC-Opt-1,X-n106-k14$^c$,14,14,1,X-n266-k58-BCR-Opt-1,X-n266-k58$^a$,61,58,1.051724138,X-n513-k21-OBC-Opt-9,X-n513-k21$^c$,21,21,1
X-n110-k13-OBC-Opt-0,X-n110-k13$^c$,13,13,1,X-n270-k35-OBC-Opt-6,X-n270-k35$^c$,36,35,1.028571429,X-n524-k153-BCR-Opt-1,X-n524-k153$^a$,160,153,1.045751634
X-n115-k10-BCR-Opt-1,X-n115-k10$^a$,10,10,1,X-n275-k28-BCR-Opt-0,X-n275-k28$^a$,31,28,1.107142857,X-n536-k96-BCR-Opt-9,X-n536-k96$^a$,98,96,1.020833333
X-n120-k6-OBC-Opt-8,X-n120-k6$^c$,6,6,1,X-n280-k17-BCR-Opt-0,X-n280-k17$^a$,17,17,1,X-n548-k50-BCR-Opt-4,X-n548-k50$^a$,55,50,1.1
X-n125-k30-BCR-Opt-9,X-n125-k30$^a$,30,30,1,X-n284-k15-OBC-Opt-8,X-n284-k15$^c$,15,15,1,X-n561-k42-BCR-Opt-1,X-n561-k42$^a$,42,42,1
X-n129-k18-OBC-Opt-4,X-n129-k18$^c$,18,18,1,X-n289-k60-BCR-Opt-1,X-n289-k60$^a$,61,60,1.016666667,X-n573-k30-OBC-Opt-2,X-n573-k30$^c$,30,30,1
X-n134-k13-OBC-Opt-0,X-n134-k13$^c$,13,13,1,X-n294-k50-BCR-Opt-6,X-n294-k50$^a$,51,50,1.02,X-n586-k159-BCR-Opt-2,X-n586-k159$^a$,171,159,1.075471698
X-n139-k10-OBC-Opt-4,X-n139-k10$^c$,10,10,1,X-n298-k31-OBC-Opt-0,X-n298-k31$^c$,32,31,1.032258065,X-n599-k92-BCR-Opt-3,X-n599-k92$^a$,95,92,1.032608696
X-n143-k7-BCR-Opt-6,X-n143-k7$^a$,7,7,1,X-n303-k21-OBC-Opt-0,X-n303-k21$^c$,21,21,1,X-n613-k62-BCR-Opt-8,X-n613-k62$^a$,62,62,1
X-n148-k46-BCR-Opt-1,X-n148-k46$^a$,49,46,1.065217391,X-n308-k13-OBC-Opt-7,X-n308-k13$^c$,13,13,1,X-n627-k43-OBC-Opt-9,X-n627-k43$^c$,44,43,1.023255814
X-n153-k22-BCR-Opt-5,X-n153-k22$^a$,23,22,1.045454545,X-n313-k71-BCR-Opt-4,X-n313-k71$^a$,73,71,1.028169014,X-n641-k35-OBC-Opt-8,X-n641-k35$^c$,35,35,1
X-n157-k13-BCR-Opt-3,X-n157-k13$^a$,15,13,1.153846154,X-n317-k53-BCR-Opt-9,X-n317-k53$^a$,64,53,1.20754717,X-n655-k131-BCR-Opt-6,X-n655-k131$^a$,164,131,1.251908397
X-n162-k11-OBC-Opt-1,X-n162-k11$^c$,11,11,1,X-n322-k28-OBC-Opt-6,X-n322-k28$^c$,28,28,1,X-n670-k130-BCR-noOpt-4,X-n670-k130$^b$,137,130,1.053846154
X-n167-k10-OBC-Opt-5,X-n167-k10$^c$,10,10,1,X-n327-k20-OBC-Opt-8,X-n327-k20$^c$,20,20,1,X-n685-k75-BCR-Opt-7,X-n685-k75$^a$,75,75,1
X-n172-k51-BCR-Opt-3,X-n172-k51$^a$,52,51,1.019607843,X-n331-k15-OBC-Opt-9,X-n331-k15$^c$,15,15,1,X-n701-k44-OBC-Opt-8,X-n701-k44$^c$,45,44,1.022727273
X-n176-k26-BCR-Opt-9,X-n176-k26$^a$,26,26,1,X-n336-k84-BCR-Opt-8,X-n336-k84$^a$,86,84,1.023809524,X-n716-k35-OBC-Opt-6,X-n716-k35$^c$,35,35,1
X-n181-k23-BCR-Opt-4,X-n181-k23$^a$,26,23,1.130434783,X-n344-k43-BCR-Opt-9,X-n344-k43$^a$,44,43,1.023255814,X-n733-k159-BCR-noOpt-2,X-n733-k159$^b$,168,159,1.056603774
X-n186-k15-OBC-Opt-6,X-n186-k15$^c$,15,15,1,X-n351-k40-OBC-Opt-6,X-n351-k40$^c$,41,40,1.025,X-n749-k98-BCR-Opt-3,X-n749-k98$^a$,99,98,1.010204082
X-n190-k8-BCR-Opt-5,X-n190-k8$^a$,8,8,1,X-n359-k29-OBC-Opt-9,X-n359-k29$^c$,30,29,1.034482759,X-n766-k71-BCR-Opt-4,X-n766-k71$^a$,72,71,1.014084507
X-n195-k51-BCR-noOpt-1,X-n195-k51$^b$,52,51,1.019607843,X-n367-k17-OBC-Opt-1,X-n367-k17$^c$,17,17,1,X-n783-k48-OBC-Opt-4,X-n783-k48$^c$,48,48,1
X-n200-k36-BCR-Opt-2,X-n200-k36$^a$,37,36,1.027777778,X-n376-k94-BCR-Opt-5,X-n376-k94$^a$,125,94,1.329787234,X-n801-k40-BCR-Opt-2,X-n801-k40$^a$,43,40,1.075
X-n204-k19-OBC-Opt-9,X-n204-k19$^c$,19,19,1,X-n384-k52-BCR-Opt-9,X-n384-k52$^a$,53,52,1.019230769,X-n819-k171-BCR-Opt-5,X-n819-k171$^a$,177,171,1.035087719
X-n209-k16-OBC-Opt-9,X-n209-k16$^c$,16,16,1,X-n393-k38-OBC-Opt-7,X-n393-k38$^c$,39,38,1.026315789,X-n837-k142-BCR-Opt-4,X-n837-k142$^a$,148,142,1.042253521
X-n214-k11-BCR-Opt-7,X-n214-k11$^a$,11,11,1,X-n401-k29-OBC-Opt-8,X-n401-k29$^c$,29,29,1,X-n856-k95-BCR-Opt-5,X-n856-k95$^a$,107,95,1.126315789
X-n219-k73-BCR-noOpt-5,X-n219-k73$^b$,109,73,1.493150685,X-n411-k19-BCR-Opt-6,X-n411-k19$^a$,19,19,1,X-n876-k59-BCR-Opt-7,X-n876-k59$^a$,59,59,1
X-n223-k34-BCR-Opt-8,X-n223-k34$^a$,35,34,1.029411765,X-n420-k130-BCR-Opt-3,X-n420-k130$^a$,140,130,1.076923077,X-n895-k37-OBC-Opt-3,X-n895-k37$^c$,38,37,1.027027027
X-n228-k23-BCR-Opt-7,X-n228-k23$^a$,23,23,1,X-n429-k61-BCR-Opt-6,X-n429-k61$^a$,62,61,1.016393443,X-n916-k207-BCR-noOpt-3,X-n916-k207$^b$,221,207,1.06763285
X-n233-k16-OBC-Opt-1,X-n233-k16$^c$,17,16,1.0625,X-n439-k37-BCR-Opt-0,X-n439-k37$^a$,40,37,1.081081081,X-n936-k151-BCR-Opt-8,X-n936-k151$^a$,157,151,1.039735099
X-n237-k14-OBC-Opt-5,X-n237-k14$^c$,14,14,1,X-n449-k29-BCR-Opt-5,X-n449-k29$^a$,29,29,1,X-n957-k87-BCR-Opt-8,X-n957-k87$^a$,96,87,1.103448276
X-n242-k48-BCR-Opt-9,X-n242-k48$^a$,50,48,1.041666667,X-n459-k26-OBC-Opt-3,X-n459-k26$^c$,26,26,1,X-n979-k58-BCR-Opt-8,X-n979-k58$^a$,58,58,1
X-n247-k50-BCR-Opt-7,X-n247-k50$^a$,51,50,1.02,X-n469-k138-BCR-Opt-7,X-n469-k138$^a$,146,138,1.057971014,X-n1001-k43-OBC-Opt-2,X-n1001-k43$^c$,43,43,1
X-n251-k28-OBC-Opt-8,X-n251-k28$^c$,28,28,1,X-n480-k70-BCR-Opt-7,X-n480-k70$^a$,72,70,1.028571429,,,,,
X-n256-k16-BCR-Opt-9,X-n256-k16$^a$,16,16,1,X-n491-k59-BCR-Opt-7,X-n491-k59$^a$,59,59,1,,,,,
\end{filecontents*}
\csvreader[separator=comma, tabular=|lcc|lcc|lcc|, table head=\hline Inst\^{}\{alg.\} &   k & k opt & Inst\^{}\{alg.\} &   k & k opt & Inst\^{}\{alg.\} &   k & k opt\\\hline, table foot=\hline]{x2.csv}{ID=\ID, Instancia=\inst,Melhor K=\avgk, K Otimo=\ok, n=\nn
ID2=\IDb, Instancia2=\instb, Melhor K2=\avgkb, K Otimo2=\okb, n2=\nb
ID3=\IDc, Instancia3=\instc, Melhor K3=\avgkc, K Otimo3=\okc, n3=\nc}
{\inst & \avgk & \ok & \instb & \avgkb & \okb & \instc & \avgkc & \okc}
} 
\end{table}

\begin{figure}[htb]
\begin{center}
\includegraphics[width=1\linewidth]{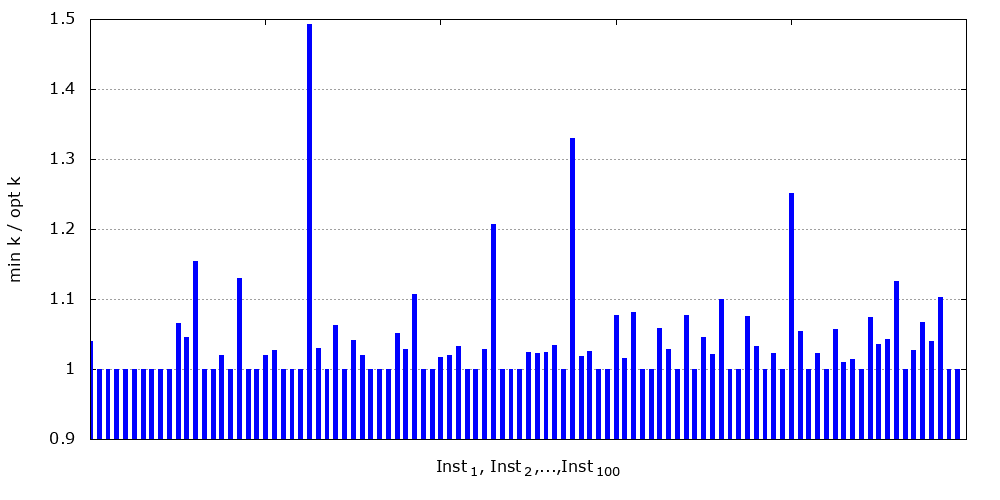}
\end{center}
\caption{Comparação do melhor $k$ obtido pelo AG em relação ao $k$ ótimo para cada uma das 100 instâncias do conjunto do X. Da esquerda para direita, cada barra corresponde a uma das instâncias, seguindo ordem crescente de $n$.}
\label{fig:x4}
\end{figure}

A Figura~\ref{fig:x4} compara o k obtido em relação ao ótimo para cada uma das 100 instâncias do conjunto X. Em 46/100 o algoritmo obteve o número de rotas ótimo.
Em 91/100  instâncias, o $k$ obtido ficou até 10\% maior que o $k$ ótimo. 

O maior tempo de execução foi de 68 minutos e o tempo médio de execução foi de 20 minutos.

\section{Conclusões}

Este trabalho propoem algoritmos para o VRP. Trabalhamos com duas variantes do problema, o PostVRP e o CVRP. No primeiro, o comprimento da rota é limitada enquanto no segudno, a capacidade do veículo é limitada.
Além disso, o PostVRP trabalha com três objetivos, número de veículos, comprimento da rota e variância entre os comprimentos das rotas enquanto o CVRP trabalha minimizando o comprimento de rota e, de maneira secundária, o número de veículos.

Em nosso trabalho simplificamos a função objetivo, trabalhando, em primeiro lugar, com número de veículos e, em segundo lugar, o comprimento da da menor rota.

Optamos por desenvolver algoritmos genéticos para resolver o problema.
O algoritmo proposto é um AG canônico com busca local. O cromossomos escolhidos não possuem divisão de rotas. No momento do \emph{cross-over}, é executado um algoritmo construtivo que faz a delimitação das rotas. Na opção \emph{LS Enable}, é executado um algoritmo de melhoria incremental da solução.

Este trabalho comparou o desempenho de um algoritmo genético usando dois operadores diferentes de \emph{crossover}, assim como a combinação do AG com busca local 2-opt. Na maioria dos casos o operador BCR obteve resultados melhores e o uso da busca local proporcionou uma convergencia mais rápida aos algoritmos. 
Ainda assim, o operador OX em conjunto ao 2-opt se mostrou uma opção viável, não só por obter resultados próximos à melhor configuração, mas também por ter um tempo de execução menor.

Em relação ao conjunto de instâncias PostVRP, mais experimentos precisam ser conduzidos, especialmente para os casos maiores a partir de 5000 pontos de entrega. Tais instâncias parecem necessitar de mais de 100  gerações. Também o algoritmo não foi testado para a maioria dos casos com mais de 10000 de entrega.

\bibliographystyle{abntex2-alf}
\bibliography{principal}

\appendix


\end{document}